\documentclass{article}

\usepackage{iclr2026_conference, times}

\usepackage[utf8]{inputenc} % allow utf-8 input
\usepackage[T1]{fontenc}    % use 8-bit T1 fonts
\usepackage{hyperref}       % hyperlinks
\usepackage{url}            % simple URL typesetting
\usepackage{booktabs}       % professional-quality tables
\usepackage{amsfonts}       % blackboard math symbols
\usepackage{nicefrac}       % compact symbols for 1/2, etc.
\usepackage{microtype}      % microtypography
\usepackage{xcolor}         % colors
\usepackage{graphicx}
 \usepackage{multirow}
\usepackage{tcolorbox}
\tcbuselibrary{skins, breakable, theorems}
\usepackage{amssymb}
\usepackage{tabularx}
\usepackage{comment}
\usepackage{algorithm}
\usepackage{algpseudocode}
\usepackage{amsmath,amssymb}
\usepackage{wrapfig}
\usepackage{enumitem}
\usepackage{makecell}
\usepackage[table]{xcolor}

\newcommand{\adr}[1]{{\color{black}#1}}  

\setlist[itemize]{leftmargin=2em}

\title{BARREL: Boundary-Aware Reasoning for Factual and Reliable LRMs
}

\author{%
  Junxiao Yang\textsuperscript{\rm 1,2}\hspace{2pt}, Jinzhe Tu\textsuperscript{\rm 1}, Haoran Liu\textsuperscript{\rm 1}, Xiaoce Wang\textsuperscript{\rm 1}, Chujie Zheng\textsuperscript{\rm 1},\\
  \textbf{Zhexin Zhang\textsuperscript{\rm 1}, Shiyao Cui\textsuperscript{\rm 1}, Caishun Chen\textsuperscript{\rm 2}, Tiantian He\textsuperscript{\rm 2}, Hongning Wang\textsuperscript{\rm 1}},\\ \textbf{Yew-Soon Ong\textsuperscript{\rm 2,3}, Minlie Huang\textsuperscript{\rm 1}\thanks{Corresponding author.}}\\
  \textsuperscript{\rm 1} The Conversational AI (CoAI) group, DCST, Tsinghua University\\
  \textsuperscript{\rm 2} Centre for Frontier AI Research, Institute of High Performance Computing, \\
  Agency for Science, Technology and Research, Singapore\\
  \textsuperscript{\rm 3} The College of Computing and Data Science, Nanyang Technological University \\
  \texttt{yangjunx21@gmail.com, aihuang@tsinghua.edu.cn} \\
}

\iclrfinalcopy

\begin{document}

\maketitle

\begin{abstract}

Recent advances in Large Reasoning Models (LRMs) have shown impressive capabilities in mathematical and logical reasoning. However, current LRMs rarely admit ignorance or respond with “I don’t know”. Instead, they often produce incorrect answers while showing undue confidence, raising concerns about their factual reliability. In this work, we identify two pathological reasoning patterns characterized by overthinking that contribute to the overconfident and incorrect answers: \emph{last-minute guessing} and \emph{second-thought spiraling}. To address these issues, we propose BARREL—a novel framework that promotes concise and boundary-aware factual reasoning. Our experiments show that BARREL-training increases the reliability of DeepSeek-R1-Distill-Llama-8B from 39.33\% to 61.48\%, while still achieving accuracy comparable to models finetuned on reasoning data generated by R1. These results demonstrate that our pilot study is inspiring to build more reliable and factual \textit{System 2} LRMs.

\end{abstract}

\section{Introduction}
\label{sec:intro}

\begin{wrapfigure}{r}{0.55\textwidth} % 
  \vspace{-1.6em}
  \includegraphics[width=0.55\textwidth]{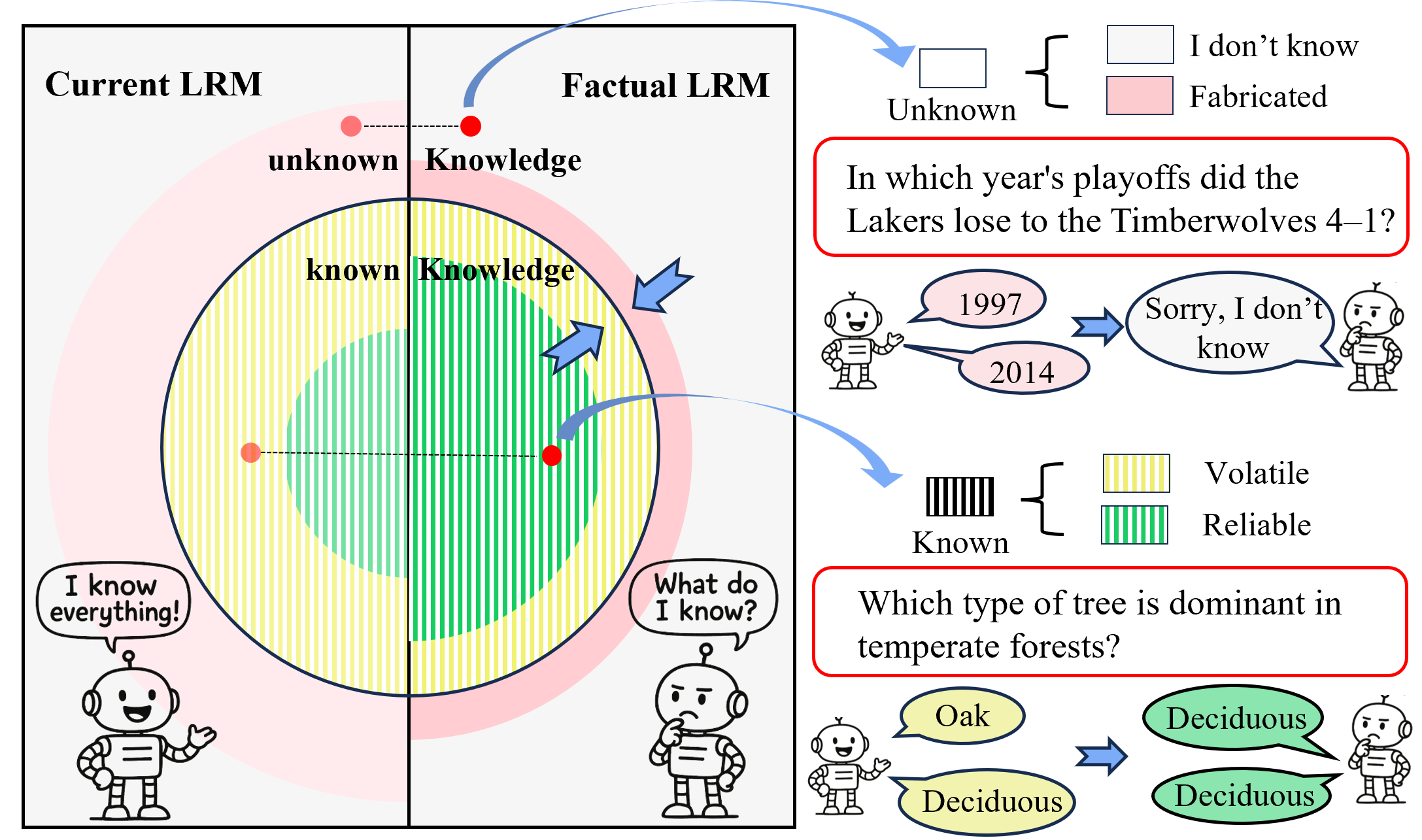}
  \vspace{-1.6em}
  \caption{How reliable factual reasoning is expected to improve model performance. \textbf{Left:} Current LRMs rarely admit ignorance and often respond inconsistently. \textbf{Right:} Reliable LRMs should acknowledge unknowns and express known facts more consistently.}
  \label{fig:intro}
  \vspace{-1.5em}
\end{wrapfigure}

Recent advances in Large Reasoning Models (LRMs) \citep{jaech2024openai, guo2025deepseek, qwq32b} have shown impressive performance in specialized reasoning tasks, especially in mathematics and logic. However, these gains have not led to corresponding improvements in reliability. On the contrary, faithfulness hallucination rates are rising \citep{HughesBae2023}, and helpfulness on factual tasks is declining \citep{zhao2025trade}, raising concerns about the reliability of these reasoning models.

In this work, we focus primarily on the factual reliability of LRMs, which is a crucial requirement for many real-world tasks. Factuality of language models involves two aspects: \textit{knowing} (whether the model holds relevant knowledge) \citep{huang2025survey, ji2023survey} and \textit{telling} (conveying the correct factual information) \citep{gekhman-etal-2024-fine, mallen2022not}. As shown in Figure \ref{fig:intro}, we want LRMs to exhibit two aspects of factual reliability: (1) Identify both what it knows and what it does not know: The model should be able to recognize questions it does not know the answer and respond with "I don't know." \citep{xu2024rejection, zhang-etal-2024-r} (2) Tell what it knows: There is a gap between knowing and telling \citep{saunders2022self}, and we want to improve the model's accuracy in expressing the knowledge it has \citep{zhang-etal-2024-self}.

However, current LRMs consistently struggle with the two factual reliability goals above. They rarely acknowledge gaps in their knowledge and often fabricate answers instead, even on questions for which they lack sufficient knowledge. Moreover, their responses can be inconsistent—providing incorrect answers in some instances while correctly responding to similar queries elsewhere \citep{wang2022self}, thereby reducing the overall factual accuracy of their responses \citep{zhao2025trade}.

\begin{wrapfigure}{l}{0.5\textwidth} % 
  \vspace{-0.9em}
  \includegraphics[width=0.5\textwidth]{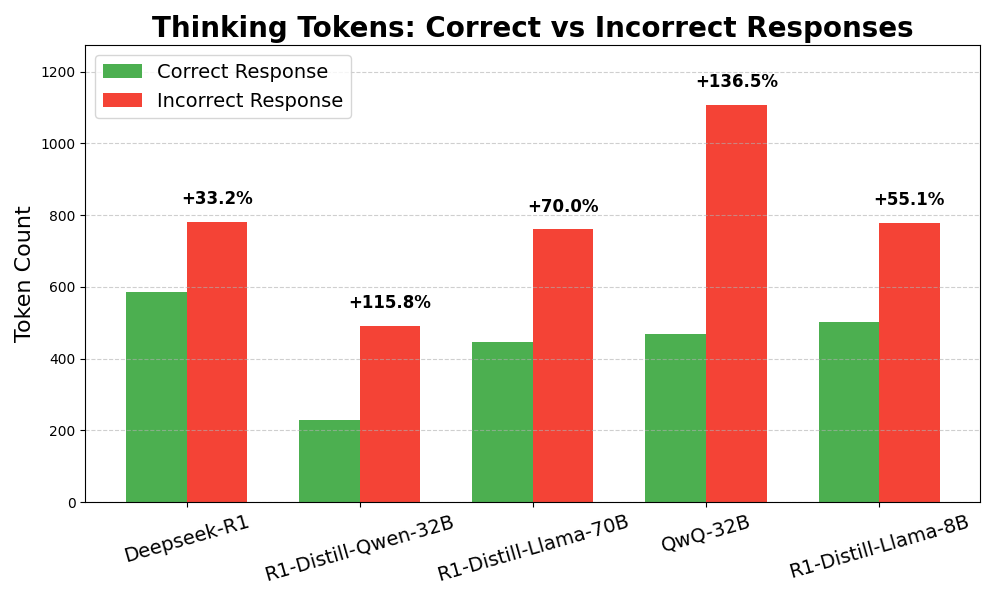}
  \vspace{-2.0em}
  \caption{Number of reasoning tokens used by LRMs when producing correct versus incorrect answers. We test on TruthfulQA across different types of reasoning models. Details and results on other datasets are listed at Appendix \ref{sec:append_pilot}.}
  \label{fig:token_number}
  \vspace{-0.6em}
\end{wrapfigure}

\begin{figure}[t]
  \centering
  \includegraphics[width=\linewidth]{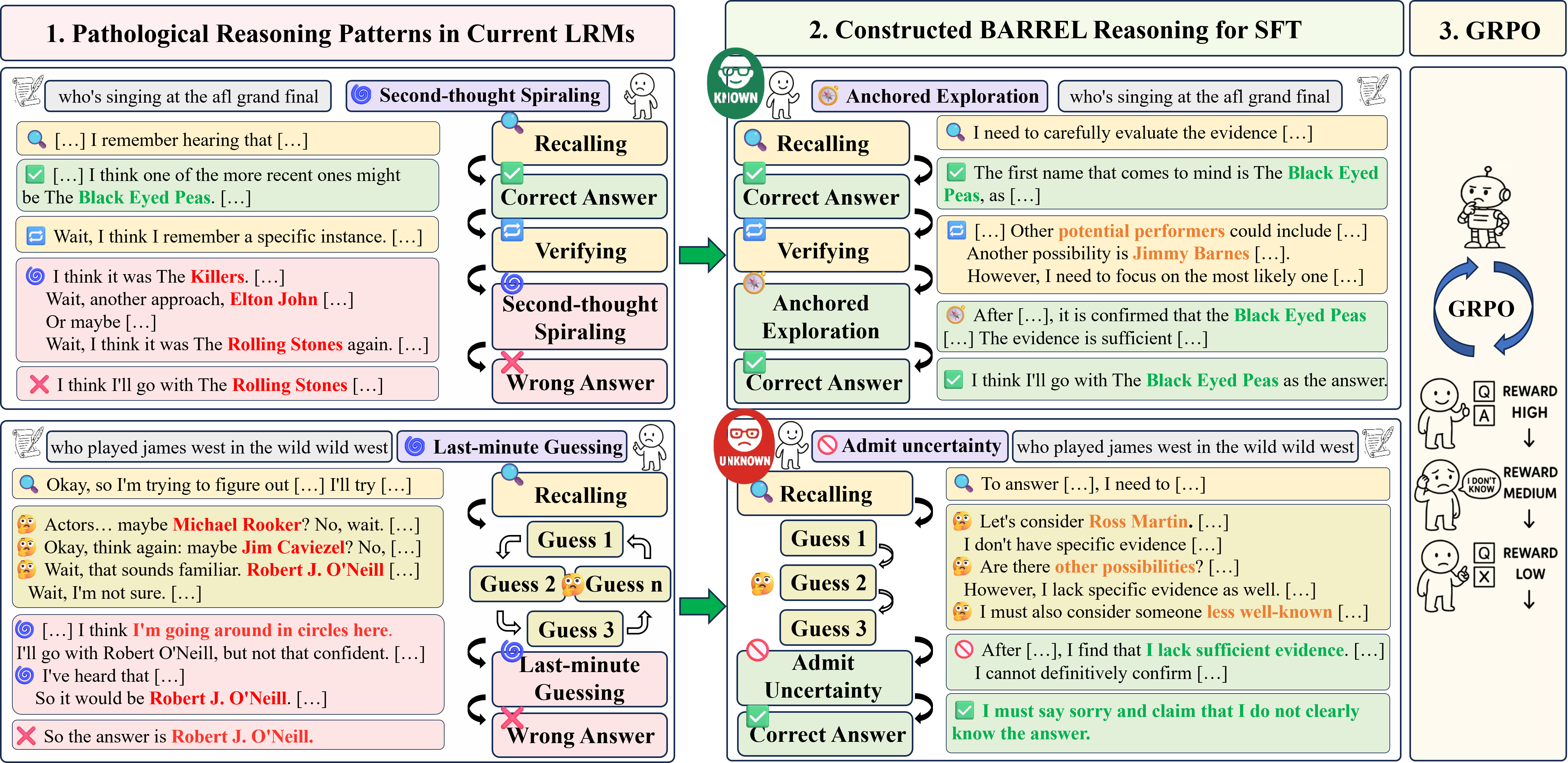}
  \vspace{-1.8em}
  \caption{
    \textbf{Left:} The two current reasoning patterns of LRM: Last-minute Guessing, typically associated with unknown knowledge, and Second-thought Spiraling, which occurs despite known knowledge. \textbf{Right:} The BARREL pipeline addresses both cases by correcting overthinking tendencies and constructing SFT data accordingly, further enhanced with GRPO.
  }
  % \vspace{-1.8em}
  \label{fig:method}
\end{figure}

What underlying reasoning behaviors contribute to these observed issues? As shown in Figure~\ref{fig:token_number}, our preliminary experiments reveal a variant of the overthinking pattern discussed by \citet{chen2024not}, which we term factual overthinking—where large reasoning models (LRMs) consume more tokens when producing factually incorrect answers than when generating correct ones. And we further identify two prevalent pathological reasoning patterns associated with this overthinking phenomenon. As shown on the left part of Figure \ref{fig:method}, one prominent pattern is \emph{\textbf{Last-minute Guessing}}, in which the model, after extensive but inconclusive reasoning, abruptly commits to an answer in a final burst of speculative output—much like a student hastily filling in an answer just before the end of an exam. Another frequent pattern is \emph{\textbf{Second-thought Spiraling}}, where the model initially identifies the correct answer but continues to over-analyze, ultimately undermining its own correct conclusion.

To mitigate these pathological reasoning patterns, we propose a novel training framework, \textbf{BARREL} (Boundary-Aware Reasoning for Reliable and Factual LRMs). As shown in Figure \ref{fig:method}, BARREL trains LRMs to perform concise, deliberative factual reasoning and draw conclusions after it has explored a sufficient number of candidates. Concretely, for known questions, when the model identifies the correct factual answer during reasoning, we want it to maintain confidence in that answer while continuing to explore other possible ones. Once there are no likely candidates, it should halt further reasoning and provide the correct factual answer. For unknown questions, after exploring a sufficient number of plausible candidates, it should terminate reasoning and proactively admit its lack of knowledge with uncertainty-aware refusal (e.g., "Sorry, I don't know"). 
To implement BARREL, we begin by employing a sampling strategy to probe the model’s knowledge boundary. Drawing on the identified pathological patterns, we construct two distinct types of reasoning data, and use Supervised Fine-Tuning (SFT) to instill the corresponding reasoning behaviors in the model. Finally, we adopt Group Relative Policy Optimization (GRPO) using general reliability-based reward (high for correct answers, medium for uncertainty-aware refusal, and low for incorrect answers) without the need for labeling \textit{known}/\textit{unknown}, further enhancing the model’s ability to generalize in factual reasoning.

Our experiments demonstrate that models trained with BARREL can effectively express uncertainty-aware refusal, and mitigate the two pathological reasoning patterns. This capability significantly improves reliability: BARREL boosts the reliability of DeepSeek-R1-Distill-Llama-8B from 39.33\% to 61.48\%, while maintaining an accuracy of 40.7\%, which is even higher than the accuracy of 38.43\% achieved by distillation. Through detailed analysis, we highlight the critical role of medium-level rewards in promoting uncertainty-aware refusal. This result also identifies the root cause of models’ inability to admit ignorance to a fundamental gap in current RL paradigms: they do not reward refusal. As a result, models are incentivized to answer every question, regardless of uncertainty. 

Our main contributions are as follows:

\begin{itemize}
    \item We discover the factual overthinking phenomenon and point out two pathological reasoning patterns that lead to factual unreliability of LRMs.
    \item To the best of our knowledge, we are the first to explore how LRMs can utilize reasoning to admit ignorance and say "I don't know". Also, we introduce a novel training pipeline to do so.
    \item We emphasize medium-level rewards to encourage uncertainty-aware refusal for reliable LRMs
\end{itemize}

% factual alignment in large language models (LLMs) has primarily employed factual supervised fine-tuning (SFT) \citep{zhang-etal-2024-r}, Direct Preference Optimization (DPO) \citep{zhang-etal-2024-self, tian2023fine} and  Proximal Policy Optimization (PPO) \citep{xu2024rejection} to instill an awareness of knowledge boundaries. Yet, these methods generally prompt models to either output "I don't know" (IDK), provide a correct answer directly—without ever teaching the model how to engage in deliberative reasoning to determine when IDK is appropriate or how to verify the correctness of its answers.

\section{Related Work}

\paragraph{Knowledge Boundary}

The knowledge boundary of LLMs refers to the extent of knowledge a model possesses or can reliably recall \citep{li2024knowledge}. Identifying such boundaries is crucial for model safety and reliability and is commonly addressed by confidence calibration \citep{ren2023investigating}, internal state probing \citep{ji2024llm}, uncertainty estimation \citep{kapoor2024large}, and sampling strategy \citep{gekhman-etal-2024-fine, xue2024ualign}. To mitigate failures by outbound queries, recent studies have proposed training LLMs to abstain from answering when uncertain, often by generating "I don't know" responses \citep{cheng2024can} or providing further explanations of the unanswerability \citep{deng2024don}. Our work pushes it further to structured and interpretable rationales about their knowledge boundary.

\paragraph{Factual Alignment}

Factual alignment aims to improve factuality while preserving instruction-following capabilities in LLMs. Recent work incorporates factuality-awareness into both SFT and RL stages to improve factual accuracy \citep{lin2024flame}, or finetuning with DPO to enhance its self-evaluation capability \citep{zhang-etal-2024-self}. Other approaches include fine-tuning with refusal-aware datasets to encourage appropriate abstention behaviors \citep{zhang-etal-2024-r}, and RLKF, which guides models to reject uncertain queries based on external feedback signals \citep{xu2024rejection}. \adr{Similarly, other reinforcement learning strategies have been proposed to incentivize models to express doubt or uncertainty \citep{stangel2025rewardingdoubtreinforcementlearning,xu2024sayselfteachingllmsexpress}. \citet{damani2025binaryrewardstraininglms} has also explored training LLMs to reason with uncertainty using binary rewards.} While previous work has primarily focused on non-reasoning models, our research demonstrates how to correct reasoning pathologies to enhance the factual reliability of LRMs.

% Simple finetuning with correct answer fails since training on unfamiliar knowledge\citep{gekhman-etal-2024-fine} may unintentionally encourage hallucinations.

% In addition to traditional metrics such as ROUGE\citep{lin2004rouge} and BLEU\citep{papineni2002bleu}, novel measures like FactScore\citep{min2023factscore} and Exact Match (EM)\citep{gekhman-etal-2024-fine} are widely adopted for evaluating factual accuracy. Furthermore, PREM, a variant of EM, has been shown to achieve higher evaluation accuracy\citep{xue2024ualign}.

% more paragraph

\section{Method}

In this section, we introduce the overall framework of \textsc{BARREL}, which comprises three main components: (1) \textbf{Knowledge Labeling}, which distinguishes whether a question is known or unknown to the model; (2) \textbf{Reasoning Trace Construction for SFT}, which constructs reasoning traces based on the question type and performs SFT to prepare the model to follow this thinking pattern; (3) \textbf{GRPO Stage}, which further enhances the model using a rule-based factual reward. We detail each component below.

\subsection{Knowledge Labeling}
\label{sec:label}

We first determine whether a question is known to the target model using the sampling strategy proposed in \citep{gekhman-etal-2024-fine}, which is a widely recognized approach \citep{xue2024ualign, li2024knowledge}. Let \(\mathcal{D}=\bigl\{(x_i,\,y_i^\ast)\bigr\}_{i=1}^{N}\) be a factual QA dataset, where each question \(x_i\) has a ground‑truth answer \(y_i^\ast\).
We generate answers with the target model \(\mathcal{M}\) using
\(K\) distinct few‑shot prompts \(\{\mathcal{P}_j\}_{j=1}^{K}\) and
repeat the sampling procedure \(L\) times for every prompt:
\[
\begin{aligned}
y_{i}^{\,j,k}\sim 
\mathcal{M}\!\left(\,\cdot \mid \mathcal{P}_{j}\,\Vert\, x_i\right),
\qquad
j=1,\dots,K,\;
k=1,\dots,L.
\end{aligned}
\tag{1}
\]
After collecting the samples $\mathcal{Y}_i \;=\;\bigl\{\,y_{i}^{\,j,k}\bigr\}_{j=1,k=1}^{K,\;L}$, we consider a question \emph{known} to the model if at least one sampled answer matches the ground-truth answer under evaluator $E$.
\[
\begin{aligned}
l_i \;=\;
\begin{cases}
\textit{known}, &\text{if } \exists\,y\in\mathcal{Y}_i\text{ such that }E(y,y_i^\ast)=1,\\[4pt]
\textit{unknown}, &\text{otherwise}.
\end{cases}
\end{aligned}
\tag{2}
\]
% Here \(l_i=1\) denotes a \textbf{known} question (parametric knowledge),
% while \(l_i=0\) denotes an \textbf{unknown} question requiring external
% knowledge.

\subsection{Reasoning Trace Construction for SFT} To address the pathological reasoning patterns identified in our analysis—namely, Last-minute Guessing and Second-thought Spiraling—we propose a targeted method for constructing reasoning trajectories. Based on the type of question, we construct two distinct evidence-grounded reasoning traces $\mathcal{T}(x_i)$ for a question $x_i$, aiming to respectively correct these two faulty reasoning patterns. This construction is outlined in Algorithm~\ref{alg:trace}.

To mitigate Second-thought Spiraling in \textbf{known} questions, where the gold answer $y^\ast$ with strong evidence $e^\ast$ is available, it should begin by retrieving and identifying this answer. It then examines alternative candidates ${(y_j, e_j)}$ to contrast possibilities. After this anchored exploration, it reaffirms the choice with solid justification and draws a confident conclusion favoring the correct answer.

To address Last-minute Guessing in \textbf{unknown} questions, the system adopts a similar exploratory strategy: it recalls background knowledge and searches on plausible answer-evidence pairs ${(y_j, e_j)}$ through hypothesizing. However, if it fails to identify a sufficiently supported answer, it explicitly acknowledges the uncertainty and ultimately outputs a cautious, confirmed rejection—demonstrating its ability to explore high-probability paths without overcommitting or hallucinating.

% \subsection{SFT-Stage}

% We begin by training the model to emulate boundary-aware and deliberative reasoning patterns using supervised fine-tuning (SFT). For each question $q$, we construct a final output $O$ by concatenating the reasoning trace $\mathcal{T}(q)$ with its corresponding answer $a^\ast$: $o = \mathcal{T}(q)\,\Vert\,a^\ast$.

% For known questions, the answer $a^\ast$ is the correct gold answer. For unknown questions, the answer is a cautious fallback (e.g., "Sorry, I don't know") reflecting the model's epistemic uncertainty. The goal of SFT is to instill the model a disciplined reasoning style grounded in traceable evidence and uncertainty-aware conclusions.

% Let $\theta$ denote the parameters of the model being trained. The supervised loss is defined as:

\begin{algorithm}
\begin{algorithmic}[1]
  \Statex \textbf{Input:} Question $x_i$ and knowledge label $l_i$, gold answer with evidence $(y^\ast,e^\ast)$, alternative candidates with poor evidence $\{(y_j,e_j)\}_{j=1}^{n}$
  \Statex \textbf{Output:} reasoning trace $\mathcal{T}(x_i)$

  \State $\mathcal{T}(x_i) \gets \langle\rangle$ \Comment{Initialize an empty trace}
  \State $\mathcal{T}(x_i) \mathrel{+}= \Call{Recall}{x_i}$  \Comment{record recalled background facts}

  \If{{$l_i = 1 (known \,\ knowledge)$}} 
      \State $\mathcal{T}(x_i) \mathrel{+}= \langle y^\ast, e^\ast \rangle$ \Comment{Attach gold answer and supporting evidence}
      \State $\mathcal{T}(x_i) \mathrel{+}= \{(y_j, e_j)\}_{j=1}^{n}$ \Comment{Attach distractor answer-evidence pairs}
      \State $\mathcal{T}(x_i) \mathrel{+}= \Call{Confirm}{y^\ast}$ \Comment{Verify the conclusion with strong evidence}
  \Else \ \ {($unknown \,\ knowledge$)}
      \State $\mathcal{T}(x_i) \mathrel{+}= \bigl\{(y_j,e_j)\bigr\}_{j=1}^{n}$ \Comment{exploring plausible answer-evidence pairs}
      \State $\mathcal{T}(x_i) \mathrel{+}= \text{Acknowledge\,\,Uncertainty}()$ \Comment{Record uncertainty for guesses}
  \EndIf

  \State \Return $\mathcal{T}(x_i)$ \Comment{Return the constructed reasoning trace}
\end{algorithmic}
\caption{BARREL reasoning trace $\mathcal{T}(x_i)$ construction}
\label{alg:trace}
\end{algorithm}

We construct the reasoning traces by prompting GPT-4 with detailed instructions and BARREL reasoning examples. This approach produces a Long-CoT-style reasoning process that aligns with the expected reasoning patterns. Examples of the constructed reasoning traces for both known and unknown questions are shown in Figure~\ref{fig:method}, and the detailed prompt used for trace construction is provided in Appendix~\ref{sec:append_prompts}.

Then, we use these data to train the model to emulate boundary-aware and deliberative reasoning patterns using SFT. For each question $x_i$, we construct full output $o_i^\ast = \mathcal{T}(x_i) \Vert a_i$, where $a_i$ is either the gold answer $y_i^\ast$ (for known questions) or an uncertainty-aware refusal (e.g., “Sorry, I don’t know”) for unknown questions. This instills the model with a disciplined reasoning style grounded in traceable evidence and uncertainty-aware conclusions. The training objective minimizes the negative log-likelihood:
\[
\begin{aligned}
\mathcal{L}(\theta)
= -\sum_{i=1}^{N}
\log P_\theta\big(o_i^\ast \mid x_i\big).
\end{aligned}
\tag{3}
\]
\subsection{GRPO-Stage}

\paragraph{Rule-Based Reward Design} 

To train the model to generate verifiable and boundary-aware reasoning trajectories and answers, we employ a rule-based reward function. We categorize the model response $o_i$ into three types, each associated with a distinct reward signal. Given a generated answer $o_i$ to question $x_i$, and ground-truth answer $y_i^\ast$, the reward function $R(o_i, y_i^\ast)$ is defined as:
\[
\begin{aligned}
R(o_i, y_i^\ast) =
\begin{cases}
r_c, & \text{if } E\,(o_i, y_i^*) = 1, \\
r_s, & \text{if } o_i \text{ contains a valid rejection phrase}, \\
r_w, & \text{otherwise}.
\end{cases}
\end{aligned}
\tag{4}
\]
This reward function provides general supervision for training the model to optimize not only for correctness but also for calibrated uncertainty, aligning with the goals of boundary-aware reasoning. It comprises three components: a high reward for a correct answer ($r_c$), a medium reward for a truthful rejection ($r_s$) and a low reward for an incorrect or hallucinated output ($r_w$). To discourage the generation of unfounded claims, the penalty for an incorrect answer is more severe than the outcome of a truthful rejection, thereby incentivizing the model to acknowledge its knowledge boundaries when uncertain. The reward magnitudes follow the order:
\[
\begin{aligned}
r_c > r_s > r_w \,.
\end{aligned}
\tag{5}
\]
% \begin{itemize}
%   \item Correct Answer Reward ($r_c$): The model provides a correct answer, encouraging accurate and verifiable outputs.
  
%   \item Incorrect Answer Penalty ($r_w$): The model produces a wrong or hallucinated answer, which is penalized more heavily than a rejection to discourage unfounded claims.
  
%   \item Truthful Rejection Reward ($r_s$): The model explicitly declines to answer when uncertain, demonstrating epistemic humility and awareness of its knowledge boundaries.
% \end{itemize}

\paragraph{GRPO Training}
After SFT, the model has learned the pattern of reasoning to express uncertainty appropriately and is able to maintain confidence when the answer is correct. Building on the above reward design, we further enhance the factual reliability of the reasoning model using Group-wise Reinforcement Policy Optimization (GRPO) \citep{shao2024deepseekmathpushinglimitsmathematical}. For each labeled input $(x_i, y^*_i, l_i)$, we sample a set of $G$ reasoning-answer trajectories from the current policy $\pi_{\theta_{\text{old}}}$:
\[
\begin{aligned}
\mathcal{O} = \left\{o_1, \dots, o_j\right\}
\sim \pi_{\theta_{\text{old}}}(\cdot \mid x_i) .
\end{aligned}
\tag{6}
\]
Each trajectory $o_j$ includes a reasoning trace followed by a final answer token. GRPO then updates the model parameters to optimize the following clipped reward-weighted objective:
\[
\begin{aligned}
    \mathcal{J}_{GRPO}(\theta) &= \mathbb{E}{[x_i \sim D, \{o_j\}_{j=1}^G \sim \pi_{\theta_{old}}(O|q)]}  \\
    & \frac{1}{G}\sum_{j=1}^G\frac{1}{|o_j|} \sum_{t=1}^{|o_i|} \left\{ \min \left[ \rho_{j,t} \hat{A}_{j,t}, \text{clip} \left( \rho_{j,t}, 1 - \epsilon, 1 + \epsilon \right)  \hat{A}_{j,t} \right] - \beta \mathbb{D}_{KL}\left[\pi_{\theta} || \pi_{ref}\right]\right\} ,
\label{eq:GRPO-obj}
\end{aligned}
\tag{7}
\]
where $\rho_{i,t} = \frac{\pi_\theta(o_{j,t} | x_i, o_{j,<t})}{\pi_{\theta_{old}}(o_{j,t} | x_i, o_{j,<t})}$ is the importance weight at step $t$, and $\mathbb{D}_{KL}$ denotes the stepwise KL divergence between the current and old policies. The advantage estimate $\hat{A}_{j,t}$ is calculated using reward normalization:
\[
\begin{aligned}
\hat{A}_{j,t} = \frac{R(o_j, y_i^*) - \bar{R}}{\sigma_r},
\qquad
\sigma_r = \sqrt{\frac{1}{G} \sum_{j=1}^{G} (R(o_j, y_i^*) - \bar{R})^2} \,.
\end{aligned}
\tag{8}
\]

\section{Experiments}
\label{sec:experiments}

\subsection{Settings}

\paragraph{Datasets}
We use separate datasets for training and evaluation. The training set consists of TriviaQA \citep{joshi-etal-2017-triviaqa}, SciQ \citep{welbl-etal-2017-crowdsourcing}, and NQ-Open \citep{kwiatkowski-etal-2019-natural}, covering general knowledge, scientific reasoning, and web-based QA, respectively. For evaluation, we sample 1,000 questions from the test splits of each dataset, forming a 3,000-question test set. 

\paragraph{Models}

Due to limited computing resource, we primarily utilize DeepSeek-R1-Distill-Llama-8B, DeepSeek-R1-Distill-Qwen-7B and Qwen3-8B to perform our study.

\paragraph{Baselines}

(1) ICL: Vanilla Reasoning models with few-shot prompt designed for factual tasks. (2) In-Context Learning with Refusal Examples (ICL-IDK): Prompting LRMs to claim uncertainty (3) Distill: SFT training using the reasoning path of DeepSeek-R1 \footnote{https://huggingface.co/deepseek-ai/DeepSeek-R1} on the training set. (4) Vanilla GRPO: A standard GRPO implementation without uncertainty-based rewards or a prior SFT stage. \adr{(5) Reliability-Enhanced GRPO: We include two variants—Vanilla GRPO w/ Verbal Confidence and Vanilla GRPO w/ Probing. These methods augment the standard GRPO by employing verbal confidence extraction and predictive classifiers, respectively, to improve reliability.} Further implementation details for all baselines are provided in Appendix \ref{sec:append_training_details}.

\paragraph{Evaluation}
We evaluate the correctness of model response by prompting the models to box their final answer as follows, and then we utilize string matching to evaluate whether the answer is correct (the model answer appears in any of the candidates). More details are listed in Appendix \ref{sec:appendix_evaluation_details}.

\begin{tcolorbox}[title = {Inference Prompt for verifiable Answer}]
Answer the following question based on your knowledge and put your final answer within \\boxed\{\}. \{question\}
\end{tcolorbox}

\paragraph{Metrics}
A Factual Reliable LRM should provide as much assistance as possible while making as few errors as possible, such that we evaluate factuality on the test set using three metrics: Accuracy (Acc.), Truthfulness (Truth.), and Reliability (Rel.) \citep{xu2024rejection}.
% $\mathcal{D} = \left\{(x_i,\,y_i^\ast)\right\}_{i=1}^{N}$ 
Let $N_c, N_r, N_w$ denote the number of correct answers, truthful rejections ("Sorry, I don't know"), and incorrect answers, respectively, where $N = N_c + N_r + N_w$. The metrics are defined as:
$$
\text{Acc.} = \frac{N_c}{N}, \quad
\text{Truth.} = \frac{N_c + N_r}{N}, \quad
\text{Rel.} = \text{ans.} \cdot \text{Truth.} + (1 - \text{ans.}) \cdot \text{Acc.}, \text{ where} \,\,\ \text{ans.} = 1 - \frac{N_r}{N}.
$$
While the Truthfulness metric (Truth.) considers the notion of truthful rejection, it overlooks the model's answer rate—since a model could achieve 100\% truthfulness simply by refusing to answer all questions. In contrast, the Reliability metric (Rel.) provides a more robust, weighted, and comprehensive evaluation by jointly considering both the truthfulness of responses and answer rate.

\paragraph{Training Details and Hyperparameters}

For BARREL-SFT and Distill SFT, we ensure that we only finetune on correct answers of the known QA set, as finetuning on unknown knowledge could encourage hallucinations \citep{gekhman-etal-2024-fine}. In practice, the rewards in GRPO stage are defined as $r_c = 1$, $r_w = -1$, and $r_s = -0.5$. More details and parameters are provided in Appendix \ref{sec:append_training_details}.

\subsection{Main Results}

% \paragraph{Balancing Accuracy and Appropriate Refusals}

\begin{table}[!h]
\centering

\renewcommand{\arraystretch}{1.5}
\resizebox{\linewidth}{!}{
\begin{tabular}{l|ccc|ccc|ccc|cccc}
\hline
\textbf{Method} & 
\multicolumn{3}{c|}{\textbf{TriviaQA}} & 
\multicolumn{3}{c|}{\textbf{SciQ}} & 
\multicolumn{3}{c|}{\textbf{NQ\_open}} & 
\multicolumn{4}{c}{\textbf{Avg.}} \\
\cline{2-14}
& \textbf{Acc. $\uparrow$} & \textbf{Truth. $\uparrow$} & \textbf{Rel. $\uparrow$}
& \textbf{Acc. $\uparrow$} & \textbf{Truth. $\uparrow$} & \textbf{Rel. $\uparrow$}
& \textbf{Acc. $\uparrow$} & \textbf{Truth. $\uparrow$} & \textbf{Rel. $\uparrow$}
& \textbf{Acc. $\uparrow$} & \textbf{Truth. $\uparrow$} & \textbf{\adr{Abstain}} & \textbf{Rel. $\uparrow$} \\
\hline
\rowcolor{cyan!5} 
\multicolumn{14}{c}{\textbf{DeepSeek-R1-Distill-Llama-8B}} \\
\hline
ICL              & 35.80  & 36.10  & 36.10  & 31.80  & 31.80  & 31.80  & 16.80  & 17.10  & 17.10  & 28.13  & 28.33  & \adr{0.20} & 28.33 \\
ICL-IDK         & 35.20  & 37.30  & 37.26  & 33.70  & 33.70  & 33.70  & 15.50  & 21.60  & 21.23  & 28.13  & 30.87  & \adr{2.74} & 30.79 \\

Distill          & 46.90  & 48.20  & 48.18  & 46.60  & 46.90  & 46.90  & 21.80  & 22.90  & 22.89  & 38.43  & 39.33  & \adr{0.90} & 39.33 \\

\adr{Vanilla GRPO} & \adr{\textbf{53.80}}  & \adr{54.30} & \adr{54.30}  & \adr{\textbf{56.80}}  & \adr{56.80}  & \adr{56.80}  & \adr{\textbf{31.10}}  & \adr{31.40}  & \adr{31.40}  & \adr{\textbf{47.23}}  & \adr{47.50}  & \adr{0.27} & \adr{47.50} \\

\adr{Vanilla GRPO w/ Verbal Conf} & \adr{45.30}  & \adr{56.40} & \adr{55.17}  & \adr{48.00}  & \adr{51.00}  & \adr{50.91}  & \adr{22.90}  & \adr{43.60}  & \adr{39.32}  & \adr{38.73}  & \adr{50.33}  & \adr{11.60} & \adr{48.99} \\

\adr{Vanilla GRPO w/ Probing} & \adr{46.20}  & \adr{60.30}  & \adr{58.31}  & \adr{51.90}  & \adr{61.50}  & \adr{60.58}  & \adr{22.80}  & \adr{54.20}  & \adr{44.34}  & \adr{40.30}  & \adr{58.67}  & \adr{18.37} & \adr{55.29} \\

BARREL              & 48.40  & \textbf{71.80}  & \textbf{66.32}  & 52.80  & \textbf{69.40}  & \textbf{66.64}  & 20.90  & \textbf{70.00}  & \textbf{45.89}  & 40.70  & \textbf{70.40}  & \adr{29.70} & \textbf{61.58} \\
\quad SFT only    & 38.10  & 55.60  & 52.54  & 39.00  & 53.50  & 51.40  & 18.50  & 40.20  & 35.49  & 31.87  & 49.77  & \adr{17.90} & 46.56 \\
\hline
\rowcolor{yellow!10} 
\multicolumn{14}{c}{\textbf{DeepSeek-R1-Distill-Qwen-7B}} \\
\hline
ICL              & 18.40  & 20.10  & 20.07  & 27.60  & 27.60  & 27.60  & 8.20   & 8.70   & 8.70   & 18.07  & 18.80  & \adr{0.73} & 18.79 \\
ICL-IDK         & 18.00  & 22.90  & 22.66  & 30.60  & 31.30  & 31.30  & 8.10   & 12.10  & 11.94  & 18.90  & 22.10  & \adr{3.20} & 22.00 \\

Distill          & 19.40  & 23.30  & 23.15  & 41.90  & 42.80  & 42.79  & 10.50  & 12.70  & 12.65  & 23.93  & 26.27  & \adr{2.34} & 26.21 \\

\adr{Vanilla GRPO} & \adr{\textbf{22.30}}  & \adr{22.30}  & \adr{22.30}  & \adr{50.00}  & \adr{50.00}  & \adr{50.00}  & \adr{\textbf{13.90}} & \adr{13.90}  & \adr{13.90}  & \adr{\textbf{28.73}}  & \adr{28.73}  & \adr{0.00} & \adr{28.73} \\

\adr{Vanilla GRPO w/ Verbal Conf} & \adr{21.40}  & \adr{21.70}  & \adr{21.70}  & \adr{38.30}  & \adr{38.30}  & \adr{38.30}  & \adr{11.80}  & \adr{12.10}  & \adr{12.10}  & \adr{23.83}  & \adr{24.03}  & \adr{0.20} & \adr{24.03} \\

\adr{Vanilla GRPO w/ Probing} & \adr{14.80}  & \adr{49.50}  & \adr{37.46} & \adr{32.40}  & \adr{66.60}  & \adr{54.90}  & \adr{6.60}  & \adr{63.40}  & \adr{31.14}  & \adr{17.93}  & \adr{59.83}  & \adr{41.90} & \adr{42.28} \\

BARREL              & 21.70  & \textbf{76.00}  & \textbf{46.52}  & \textbf{50.60}  & \textbf{64.20}  & \textbf{62.35}  & 12.50  & \textbf{83.30}  & \textbf{33.17}  & 28.27  & \textbf{74.50}  & \adr{46.23} & \textbf{53.12} \\
\quad SFT only    & 17.00  & 38.90  & 34.10  & 34.60  & 43.90  & 43.04  & 10.00  & 33.70  & 28.08  & 20.53  & 38.83  & \adr{18.30} & 35.48 \\
\hline

\rowcolor{orange!10}
\multicolumn{14}{c}{\textbf{Qwen3-8B}} \\
\hline
ICL              & 50.20 & 51.00 & 50.99 & 52.60 & 52.60 & 52.60 & 23.10 & 23.60 & 23.60 & 41.97 & 42.40 & \adr{0.43} & 42.40 \\
ICL-IDK          & 51.10 & 55.10 & 55.40 & 54.90 & 55.30 & 55.30 & 23.90 & 34.10 & 33.06 & 43.30 & 48.17 & \adr{4.87} & 47.93 \\
Distill          & 52.90 & 54.60 & 54.67 & 57.00 & 57.20 & 57.20 & 24.80 & 26.20 & 26.18 & 44.90 & 46.00 & \adr{1.10} & 45.99 \\
\adr{Vanilla GRPO} & \adr{54.50}  & \adr{54.90}  & \adr{54.90}  & \adr{63.50}  & \adr{63.50}  & \adr{63.50}  & \adr{\textbf{33.80}} & \adr{33.90}  & \adr{33.90}  & \adr{\textbf{50.60}}  & \adr{50.77}  & \adr{0.17} & \adr{50.77} \\

\adr{Vanilla GRPO w/ Verbal Conf} & \adr{52.40}  & \adr{52.60}  & \adr{52.60}  & \adr{63.10}  & \adr{63.10}  & \adr{63.10}  & \adr{31.40}  & \adr{31.40}  & \adr{31.40}  & \adr{48.97}  & \adr{49.03}  & \adr{0.06} & \adr{49.03} \\

\adr{Vanilla GRPO w/ Probing} & \adr{45.80}  & \adr{63.00}  & \adr{60.04}  & \adr{58.20}  & \adr{66.80}  & \adr{66.06}  & \adr{20.90}  & \adr{61.90}  & \adr{45.09}  & \adr{41.63}  & \adr{63.90}  & \adr{22.27} & \adr{58.94} \\
BARREL            & \textbf{55.50} & \textbf{86.50} & \textbf{76.89} & \textbf{69.30} & \textbf{79.10} & \textbf{78.14} & 26.70 & \textbf{75.60} & \textbf{51.79} & 50.50 & \textbf{80.40} & \adr{29.90} & \textbf{71.46} \\
\quad SFT only    & 40.90 & 57.00 & 54.41 & 52.50 & 65.00 & 63.44 & 19.60 & 36.60 & 33.71 & 37.67 & 52.87 & \adr{15.20} & 50.56 \\
\hline

\hline
\end{tabular}
}
\vspace{-0.0em}
\caption{Comparison of Different Methods on Accuracy, Truthfulness, and Reliability Across Datasets. All results are multipled by 100.}
\vspace{-0.8em}
\label{tab:main_res}
\end{table}

\paragraph{Balancing Accuracy and Appropriate Refusals} Our experimental results in Table \ref{tab:main_res} demonstrate that our method significantly enhances model reliability and truthfulness, while maintaining accuracy. For the baseline methods, the truthfulness and reliability scores consistently remain below 40\%. These models rarely acknowledge uncertainty. In contrast, our method increases the reliability of DeepSeek-R1-Distill-Llama-8B from 39.33\% to 61.48\%, while maintaining an accuracy of 40.7\%, surpassing the distillation method's 38.43\%. Similar improvements are observed for the DeepSeek-R1-Distill-Qwen-7B and Qwen3-8B. \adr{We further compare BARREL with Vanilla GRPO and post-hoc confidence estimation methods (Verbal Confidence and Probing). While Vanilla GRPO lacks the mechanism to express uncertainty, and probing methods often suffer from accuracy degradation due to miscalibration when tuned for higher truthfulness (Detailed in Appendix \ref{sec:addtional_baselines}), BARREL consistently yields a superior balance for reliability. This highlights the advantage of using RL to teach LRMs to reason and internalize the accuracy-refusal trade-off, rather than relying on external classifiers or heuristic thresholds. For instance, on Qwen3-8B, BARREL achieves a reliability score of 71.46\%, significantly surpassing the Probing baseline of 58.94\%.} Table \ref{tab:main_example} provides examples illustrating how BARREL-trained LRMs mitigate Last-minute Guessing and Second-thought Spiraling. Overall, our training method enables LRMs to retain relatively high accuracy while expressing uncertainty on approximately 50\% of the remaining questions, thereby substantially improving factual reliability. 

\paragraph{Discussion on the Two Stages of BARREL} 

\begin{wrapfigure}{r}{0.5\textwidth} % 
  \vspace{-1.8em}
  \includegraphics[width=0.5\textwidth]{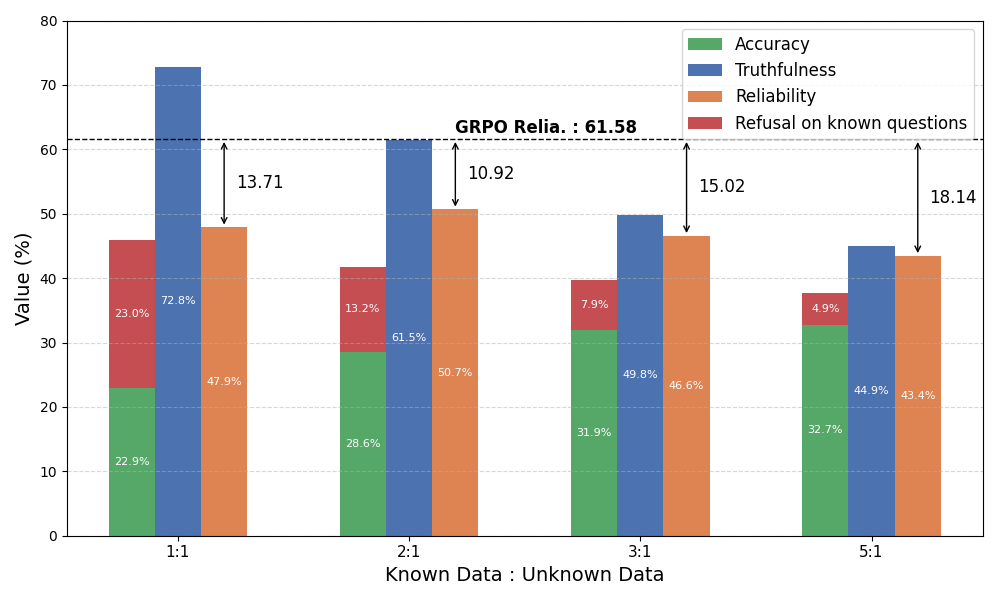}
  \vspace{-2.2em}
  \caption{Effect of the ratio of \textit{known data : unknown data} on the factuality scores of SFT models.}
  \label{fig:sft}
  \vspace{-0.8em}
\end{wrapfigure}
We can notice from the results in Table \ref{tab:main_res} that GRPO training is indeed necessary. Although the SFT model performs well in terms of truthfulness, its accuracy remains relatively low. The SFT process primarily helps the model learn basic refusal patterns, but its effectiveness is limited—we discuss this in more detail in Section \ref{sec:analysis_sft_trade_off}. Table \ref{tab:main_example} presents several examples showing that GRPO can address the issues caused by SFT, such as excessive refusals and faulty reasoning. By providing general supervision signals, GRPO enables the model to self-adjust (instead of relying on external knowledge injection). This further improves both accuracy and truthfulness, making the model more factually reliable. \adr{Results also demonstrate that BARREL SFT effectively mitigates the ``factual overthinking'' in Figure \ref{fig:token_number} and that GRPO further promotes concise, reliable reasoning without sacrificing accuracy. We provide a detailed analysis in Appendix~\ref{app:response_length}.} We also find that only conducting GRPO training with the uncertainty reward is insufficient for teaching LRMs to be reliable. We discuss this ablation in Appendix \ref{sec:append_grpo}.

% \paragraph{Comparison to GRPO and GRPO}

\subsection{Analysis}

\paragraph{Trade-off in SFT Stage}
\label{sec:analysis_sft_trade_off}

We found that teaching refusal during SFT introduces a clear trade-off: the SFT model tends to exhibit either low accuracy or low truthfulness, making it difficult to achieve both. As shown in Figure \ref{fig:sft}, as the ratio between \textit{known data} and \textit{unknown data} increases, the model's accuracy improves, but its truthfulness declines significantly. This ultimately imposes a ceiling on the model's reliability, leaving a noticeable gap compared to results obtained through GRPO. It is worth noting that the accuracy improvement caused by increasing the ratio does not reflect an actual improvement in model capability—it merely reduces the number of incorrect refusals on known questions. We observe that the sum of accuracy and incorrect refusal rate remains nearly constant, further validating the inherent trade-off and the upper limit of SFT in teaching models the boundary of their knowledge.

\begin{table}[!t]
    % \small
    \centering
    \resizebox{\linewidth}{!}{
    \begin{tabular}{|p{4.8cm}|p{5cm}|p{5.2cm}|p{5cm}|}
    \toprule
    \multicolumn{4}{c}{\textbf{Reasoning Examples for different models}} \\
    
    \midrule
    \textbf{ICL-IDK} & \textbf{Distill} & \textbf{Ours (SFT)} & \textbf{Ours (GRPO)}\\ 
    \midrule
    
    \multicolumn{4}{|c|}{Question: On which island is Reina Sofia Airport?} \\
    \midrule
    \textbf{Answer: \textcolor{red!80!black}{Crete}} \par \textbf{Reasoning: }Okay, so I need to figure out ... \textcolor{orange!75!black}{Hmm, I'm not entirely sure about this one.} ... Let me try to recall any other places with Reina Sofia. I don't think ... \textcolor{red!80!black}{Maybe it's on Crete.} \textcolor{green!60!black}{I should also consider other possibilities.} ... \textcolor{red!80!black}{I don't think so} because the name Reina Sofia doesn't ring a bell with those regions. ... \textcolor{orange!75!black}{I'm a bit confused because} ... Wait, I think I've heard that ... \textcolor{red!80!black}{So, the island would be Crete.} Yeah, that makes sense. So the answer is Crete. & 

    \textbf{Answer: \textcolor{green!60!black}{Tenerife}} \par \textbf{Reasoning: }Okay, so the question is ... I think it's a \textcolor{green!60!black}{major airport}, maybe in a \textcolor{green!60!black}{European country}. The name Reina Sofia sounds \textcolor{green!60!black}{Spanish} to me. ... Reina Sofia ... {Wait, Reina Sofia might be in Tenerife.} Because ... maybe their main airport is called \textcolor{green!60!black}{Reina Sofia.} Let me check. \textcolor{green!60!black}{Yes, I think the main airport in Tenerife is named Reina Sofia.} The capital of the Canary Islands is Las Palmas, \textcolor{red!80!black}{but the largest island by population is Tenerife.} So putting it all together, the answer should be Tenerife. & 
    
    \textbf{Answer: \textcolor{red!80!black}{Gran Canaria}} \par \textbf{Reasoning: }To determine on which island ...  \textcolor{green!60!black}{I should consider other islands in the Canary Islands}, such as ... However, Tenerife has its own major airport ... \textcolor{red!80!black}{so they are unlikely candidates.} ... \textcolor{green!60!black}{Next, I should question} whether there could be any confusion with other Spanish-speaking islands ... However ...  \textcolor{green!60!black}{To verify}, I recall that Reina Sofia Airport ... \textcolor{red!80!black}{international tourism in the Canary Islands, and Gran Canaria is the most popular ...} conclusion is that Reina Sofia Airport is on Gran Canaria. & 
    
    \textbf{Answer: \textcolor{green!60!black}{Tenerife}} \par \textbf{Reasoning: }To determine on which island ... \textcolor{green!60!black}{The most prominent island ... Tenerife}, which is often \textcolor{green!60!black}{associated with Reina Sofia Airport.} However, \textcolor{green!60!black}{I need to verify this} by considering ... 1. \textcolor{green!60!black}{**Tenerife**}: ... The evidence supporting Tenerife includes the fact that Reina Sofia Airport is \textcolor{green!60!black}{frequently linked to Tenerife} ... 2. **Gran Canaria**: ... 3. **Fuerteventura**: ... Now, I need to \textcolor{green!60!black}{critically evaluate} the evidence for Tenerife. ... After considering the possibilities ...  Tenerife. \\
    
    \midrule
    
    \multicolumn{4}{|c|}{Question: What colour is the liqueur ‘Parfait d’Amour’?} \\
    \midrule
    \textbf{Answer: \textcolor{red!80!black}{golden}}
    
    \textbf{Reasoning: }Okay, so I need to figure out ... Wait, ... \textcolor{orange!75!black}{maybe} with a \textcolor{red!80!black}{golden color}. \textcolor{orange!75!black}{Or was it} more of a yellowish hue? \textcolor{orange!75!black}{I'm a bit confused.} ... but I think Parfait d'Amour is more \textcolor{red!80!black}{on the golden side.} ... \textcolor{orange!75!black}{Maybe} it's a clear, golden color because of the ingredients used. ... Another angle: sometimes liqueurs are aged ... which could give it a \textcolor{red!80!black}{amber or golden hue.} ... \textcolor{orange!75!black}{I'm still not sure.} ... \textcolor{orange!75!black}{I think it's more likely} to be a warm color like \textcolor{red!80!black}{amber or golden.} I'm going to go with that, but \textcolor{orange!75!black}{I'm not 100\% sure.} I might have to look it up later, but for now, I'll say it's a clear, \textcolor{red!80!black}{golden color.} & 

    \textbf{Answer: \textcolor{red!80!black}{white}}
    
    \textbf{Reasoning: }Okay, so I need to figure out ... I recall that Parfait d'Amour is a French liqueur \textcolor{red!80!black}{made with chocolate, vanilla,} and ... \textcolor{orange!75!black}{Maybe} Parfait d'Amour is a \textcolor{red!80!black}{white chocolate} liqueur? \textcolor{orange!75!black}{Or maybe} it's a \textcolor{red!80!black}{milk chocolate} one. Alternatively, \textcolor{orange!75!black}{maybe} it's a \textcolor{red!80!black}{cream} liqueur. Let me think. Parfait d'Amour might be a \textcolor{red!80!black}{white} liqueur ... Alternatively, \textcolor{orange!75!black}{maybe} it's a coffee liqueur. \textcolor{orange!75!black}{No, but I think} Parfait d'Amour is specifically \textcolor{red!80!black}{chocolate}. ...  \textcolor{orange!75!black}{Wait} ... I think I've heard that it's \textcolor{red!80!black}{white}. \textcolor{green!60!black}{Let me confirm.} ... Wait, but maybe the answer is white. \textcolor{green!60!black}{Let me check my memory again.} ... So the color is white. & 
    
    \textbf{Answer: \textcolor{orange!75!black}{I don't know}} 
    
    \textbf{Reasoning: }The user is asking about ... To answer this, \textcolor{green!60!black}{I need to consider the possibilities } ... One possibility is that it is clear. ... \textcolor{green!60!black}{However, I need to ask:} ... \textcolor{orange!75!black}{No, I don't have direct evidence that it is clear.} ... Another possibility is that ... \textcolor{green!60!black}{But again, what evidence supports this?} ... \textcolor{orange!75!black}{No, I don't have any specific evidence to support this claim either.} ...A third possibility is ... I have considered multiple possibilities ... and critically examined each one. \textcolor{green!60!black}{For each, I have asked myself what specific evidence supports the claim} and found that \textcolor{orange!75!black}{I lack sufficient information to confidently determine the color of 'Parfait d'Amour'.} & 
    
    \textbf{Answer: \textcolor{orange!75!black}{I don't know}} 
    
    \textbf{Reasoning: }The user asks me about ... \textcolor{green!60!black}{I need to consider the possible colors that liqueurs can come in} ... 1. Red: \textcolor{green!60!black}{What specific evidence supports} ... red? ... \textcolor{orange!75!black}{I don't have evidence that this particular liqueur is red.} 2. Blue: ... \textcolor{orange!75!black}{However, I don't recall any specific evidence} connecting 'Parfait d'Amour' to blue as its color. 3. Green: Green is another possibility, ... \textcolor{orange!75!black}{However, I don't have any specific evidence} that 'Parfait d'Amour' is green. ... After considering these possibilities, \textcolor{orange!75!black}{I find that I don't have sufficient evidence} ... I cannot confirm this with the information I have. \\
    \midrule
    
    \multicolumn{4}{|c|}{Question: Who composed the opera Turandot?} \\
    \midrule
    \textbf{Answer: \textcolor{red!80!black}{Sergei Prokofiev}}
    
    \textbf{Reasoning: }Okay, so I have to figure out who composed the opera Turandot. \textcolor{green!60!black}{I remember that Puccini is a famous composer,} \textcolor{orange!75!black}{but I don't think he did Turandot.} Wait, wasn't there another composer? \textcolor{red!80!black}{Oh right, Sergei Prokofiev!} He composed several operas, and \textcolor{red!80!black}{I think Turandot is one of them.} \textcolor{green!60!black}{But wait, wasn't there another version} by someone else? \textcolor{orange!75!black}{I'm a bit confused now.} ... \textcolor{red!80!black}{No, wait, I think Prokofiev's is the most famous one.} So I'm pretty confident the composer is Sergei Prokofiev. I should double-check, but I think that's right.& 

    \textbf{Answer: \textcolor{red!80!black}{Giuseppe Verdi}}
    
    \textbf{Reasoning: }Okay, so I need to figure out ... \textcolor{red!80!black}{Wait, Giuseppe Verdi comes to mind.} He was a big composer ... I remember that \textcolor{red!80!black}{Turandot is one of his later works.} Let me think. ... \textcolor{green!60!black}{But wait, I should make sure I'm not confusing it with someone else.} Let me think of other Italian composers. There's \textcolor{orange!75!black}{Puccini, but ... Then there's Rossini, but he was earlier.} ... So Verdi seems right. \textcolor{green!60!black}{Another way to check:} Turandot's plot involves ... which matches Verdi's style. ... \textcolor{red!80!black}{Wait, Verdi's last opera was Falstaff, which was in 1894, ...} So the answer should be Giuseppe Verdi. ... & 
    
    \textbf{Answer: \textcolor{orange!75!black}{I don't know}}
    
    \textbf{Reasoning: }The user asks me about ... I need to ... \textcolor{green!60!black}{evaluate whether I have sufficient evidence} to support any specific claim. .. \textcolor{orange!75!black}{Could the composer be Mozart?} ... \textcolor{green!60!black}{What evidence supports this?} I don't have any recollection of ... so I must reject this possibility. Could the composer be Beethoven? ... \textcolor{orange!75!black}{What evidence supports this?} ... Could the composer be Puccini? ... \textcolor{orange!75!black}{I cannot definitively confirm} that he composed Turandot without more concrete evidence. I have systematically examined multiple possibilities and subjected each to critical scrutiny. \textcolor{orange!75!black}{For each candidate, ... I lack sufficient evidence ...} & 
    
    \textbf{Answer: \textcolor{green!60!black}{Giacomo Puccini}}
    
    \textbf{Reasoning: }To determine ... I need to ... The name that comes to mind is \textcolor{green!60!black}{Giacomo Puccini,} ... However, I need to \textcolor{green!60!black}{verify this carefully.} ... Turandot is often linked to Puccini, ... \textcolor{green!60!black}{However, I should consider other} possible composers ... \textcolor{orange!75!black}{Another possibility could be Giuseppe Verdi,} ... However, \textcolor{green!60!black}{I do not recall him being linked to Turandot.} \textcolor{orange!75!black}{I should also consider other composers} ... However, \textcolor{green!60!black}{I do not recall him} being connected to Turandot. To further verify, \textcolor{green!60!black}{I need to examine the evidence.} ... After ... \textcolor{green!60!black}{finding no evidence to support them,} I conclude that ... Giacomo Puccini. \\
    \midrule
    
    \bottomrule
    \end{tabular}
    }
    \vspace{-1ex}
    \caption{Reasoning path examples of different methods on DeepSeek-R1-Distill-Llama-8B. We colored with \textcolor{red!80!black}{wrong factual thinking leading to \textbf{wrong answers}}, \textcolor{orange!75!black}{factual thinking alerting \textbf{not sure}} and the \textcolor{green!60!black}{good factual thinking leading to \textbf{correct answers}}.}
    \label{tab:main_example}
    \vspace{-3ex}
\end{table}

\begin{figure}[H]
  \centering
  \includegraphics[width=0.9\linewidth]{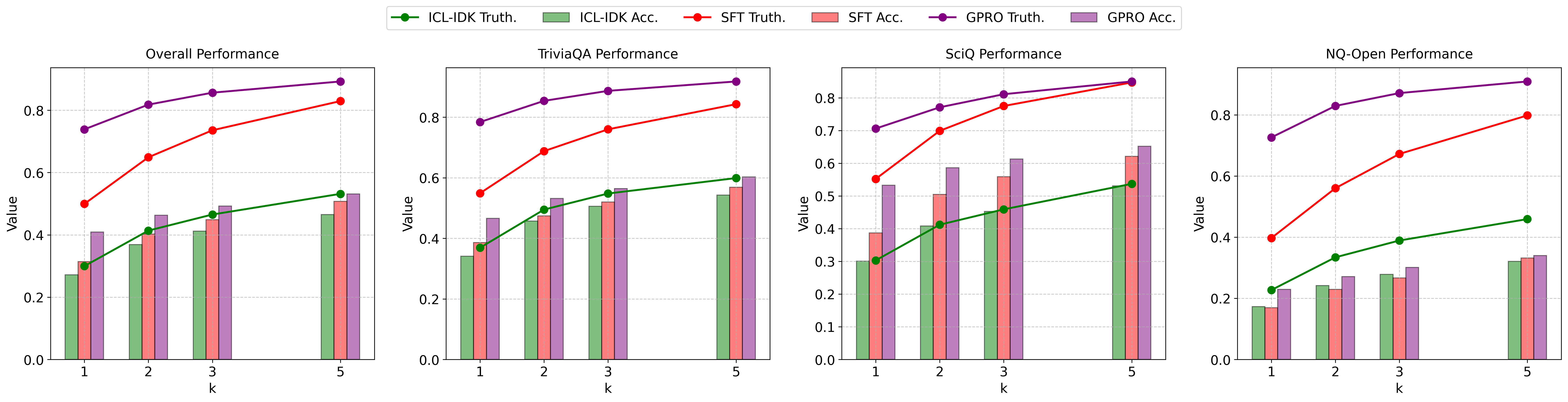}
  \vspace{-1.8em}
  \caption{
    The pass@k accuracy and truthfulness score on DeepSeek-R1-Distill-Llama-8B. We show the similar results on DeepSeek-R1-Distill-Qwen-7B at Appendix \ref{sec:append_pass_k}.
  }
  \vspace{-0.8em}
  \label{fig:pass_k_res}
\end{figure}

\begin{wrapfigure}{r}{0.5\textwidth} % 
  \vspace{-0.8em}
  \includegraphics[width=0.5\textwidth]{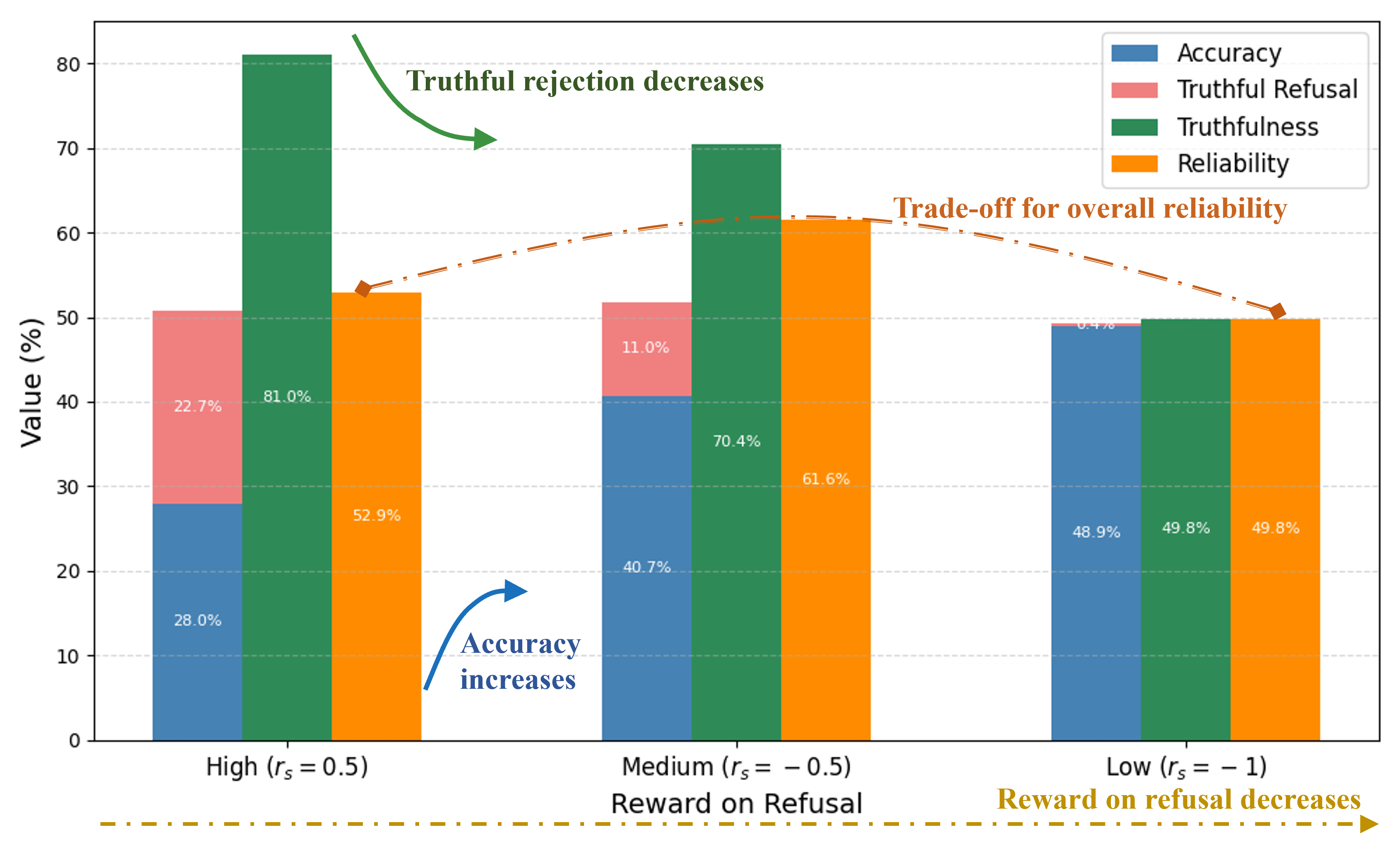}
  \vspace{-2.2em}
  \caption{Effect of the reward on refusal on the factuality scores of GRPO models.}
  \label{fig:grpo}
  \vspace{-1.1em}
\end{wrapfigure}
\paragraph{Determining Appropriate Reward for Truthful Rejection}
\label{sec:grpo_ablation}
We also explored whether the reward for refusal responses, $r_s$, is necessary in GRPO and what role it actually plays. As shown in Figure \ref{fig:grpo}, removing $r_s$—that is, treating the reward for saying "I don't know" the same as for incorrect answers—results in the model almost never admitting uncertainty or refuses to answer, even when training starts from an SFT model that already has refusal patterns. This effectively explains why existing LLMs exhibit this pattern: on one hand, we haven't taught the model how to reason about its knowledge boundaries and proactively acknowledge them; on the other hand, current RL training does not reward refusal, thus forcing the model to adopt a strategy of attempting to answer regardless. We also found that setting the reward too high leads to an excessive rate of refusal, similar to the behaviors observed in SFT.

\begin{wrapfigure}{l}{0.5\textwidth} % 
  \vspace{-1.2em}
  \includegraphics[width=0.5\textwidth]{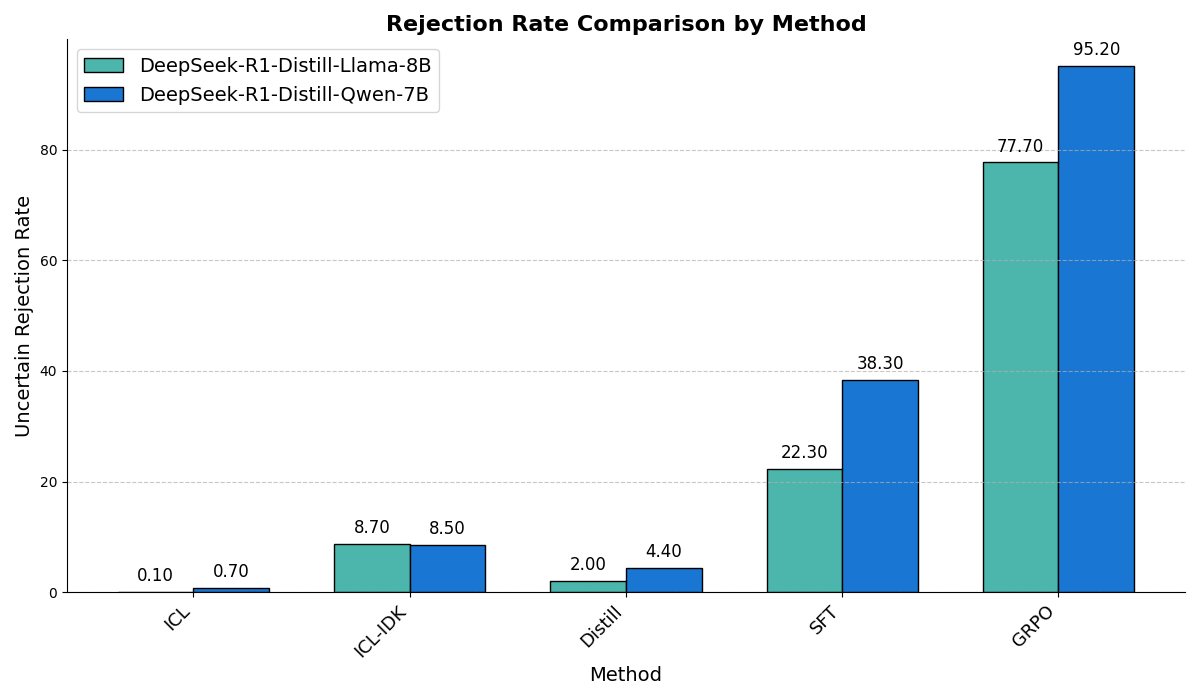}
  \vspace{-1.9em}
  \caption{Uncertain Refusal Rate on OOD unknown dataset, conducted on DeepSeek-R1-Distill-Qwen-7B and DeepSeek-R1-Distill-Llama-8B.}
  \label{fig:simpleqa}
  \vspace{-1.2em}
\end{wrapfigure}

\paragraph{Refusal Rate on OOD unknown Dataset} We additionally sample 1,000 questions from the complex SimpleQA test set \citep{wei2024measuring}. Since both models have an accuracy of around 3.0\%, we treat this dataset as unknown and use it to assess the uncertain refusal performance on an almost unknown OOD dataset. As shown in Figure \ref{fig:simpleqa}, the uncertain refusal ability of BARREL trained models could generalize to an OOD unknown dataset, saying "Sorry, I don't know" on most occasions. We also test on the unanswerable questions \citep{yin-etal-2023-large} and find that BARREL-trained models refuse over 96\% of them, as listed in Appendix \ref{sec:append_unanswerable}.

\paragraph{Did GRPO sacrifices the potential of pass@k in exchange for pass@1 performance?}

Recent studies \citep{yue2025does} have pointed out that reasoning models trained with RL may only improve performance at pass@1. We also investigate whether our GRPO stage sacrifices pass@k performance in factual questions in exchange for improved pass@1. \adr{In this context, we specifically examine whether the absolute pass@k performance of GRPO falls below the SFT baseline.} As shown in Figure \ref{fig:pass_k_res}, we observe that as k increases, the original reasoning model, the SFT model, and the GRPO model follow a similar trend as pass@1. The GRPO model’s pass@5 accuracy and truthfulness remain higher than those of baseline methods, indicating GRPO does not sacrifice pass@k performance for better pass@1 results.

\begin{wraptable}{r}{0.4\textwidth}
\vspace{-0.3em}
\centering
\renewcommand{\arraystretch}{1.4}
\resizebox{0.4\textwidth}{!}{
\begin{tabular}{l|ccc}
\hline
\multirow{2}{*}{\textbf{Method}} & \multicolumn{3}{c}{\textbf{Factual Avg.}} \\
\cline{2-4}
& \textbf{Acc.} & \textbf{Truth.} & \textbf{Rel.} \\
\hline
ICL-IDK & 28.13 & 30.87 & 30.79 \\
\hline
\multicolumn{4}{c}{\textbf{Strategy: Constructing (Default)}} \\
\hline
BARREL (SFT Only) & 31.87 & 49.77 & 46.56 \\
BARREL (Full) & 40.70 & 70.40 & 61.58 \\
\hline
\multicolumn{4}{c}{\textbf{Strategy: Rewriting}} \\
\hline
BARREL (SFT Only) & 27.03 & 44.13 & 41.21 \\
BARREL (Full) & \textbf{41.20} & \textbf{73.80} & \textbf{63.17} \\
\hline
\end{tabular}
}
\vspace{-0.2em}
\caption{Results for Constructing vs. Rewriting strategies (DeepSeek-R1-Distill-Llama-8B).}
\label{tab:ablation_strategies}
\vspace{-1.0em}
\end{wraptable}

\paragraph{\adr{Ablation on SFT data construction: Constructing vs. Rewriting}}

To validate the generalizability of our data synthesis pipeline, we conducted an ablation study comparing our default \textit{Constructing} strategy (generating traces from scratch via GPT-4) against a \textit{Rewriting} strategy. In the latter approach, we utilize GPT-4 to revise failed trajectories generated by the student model into the BARREL format. We performed this comparison using DeepSeek-R1-Distill-Llama-8B. As shown in Table~\ref{tab:ablation_strategies}, the Constructing strategy yields superior performance during the SFT stage, outperforming the Rewriting approach by approximately 5 points across metrics. We attribute this to the inherent challenge of correcting low-quality traces from a smaller-scale model, where generating high-quality reasoning from scratch proves more effective initially. However, after applying the GRPO stage, the performance gap closes, with the Rewriting strategy achieving comparable—and slightly superior—results (e.g., 63.17\% Reliability vs. 61.58\%). This indicates that the GRPO stage effectively mitigates initial SFT data discrepancies, suggesting that refining real-world failure cases is a viable and scalable alternative for BARREL framework.
% \paragraph{From implict awareness to explict awareness} 

\paragraph{Will uncertainty refusal influence math reasoning ability?}

\begin{wraptable}{r}{0.5\textwidth} % r 表示表格靠右，0.6\textwidth 控制表格宽度
\vspace{-1.2em}
\centering
\renewcommand{\arraystretch}{1.4}
\resizebox{0.5\textwidth}{!}{ % 缩小表格（略小于 wraptable 的宽度）
\begin{tabular}{l|ccc|c}
\hline
\textbf{Method} & 
\multicolumn{3}{c|}{\textbf{Factual Avg.}} & 
\textbf{MATH500} \\
\cline{2-5}
& \textbf{Acc. $\uparrow$} & \textbf{Truth. $\uparrow$} & \textbf{Rel. $\uparrow$} & \textbf{Acc. $\uparrow$} \\
\hline
\multicolumn{5}{c}{\textbf{DeepSeek-R1-Distill-Llama-8B}} \\
\hline
Original Model   & 28.13 & 30.87 & 30.79 & 81.80 \\
BARREL Trained   & 40.90 & 72.97 & 62.68 & 81.00\\
\hline
\multicolumn{5}{c}{\textbf{DeepSeek-R1-Distill-Qwen-7B}} \\
\hline
Original Model   & 8.90 & 22.10 & 22.00 & 85.60 \\
BARREL Trained   & 28.27 & 74.50 & 53.12 & 86.80 \\
\hline
\end{tabular}
}
\caption{Results on Math500 test set.}
\label{tab:math}
\vspace{-1.8em}
\end{wraptable}

As shown in Table \ref{tab:math}, we conducted additional experiments on the MATH500\footnote{https://huggingface.co/datasets/HuggingFaceH4/MATH-500} test set and found that models trained with BARREL exhibit comparable mathematical reasoning performance. For Distill-Llama-8B, we included a subset of MATH \citep{DBLP:conf/nips/HendrycksBKABTS21} in the training data. These results validate that incorporating uncertainty-based refusal does not compromise the mathematical reasoning capabilities of LRMs.

% \section{discussion}

% \subsection{Did GRPO sacrifices the potential of pass@k in exchange for pass@1 performance?}

% Recent studies have 

% \begin{figure}[H]
%   \centering
%   \includegraphics[width=\linewidth]{figs/passk_results.png}
%   \caption{
%     The pass@k accuracy and truthfulness score on DeepSeek-R1-Distill-Llama-8B. We show the similar results on DeepSeek-R1-Distill-Qwen-7B at Appendix \ref{sec:append_pass_k}
%   }
%   \label{fig:pass_k_res}
% \end{figure}

\section{Conclusion}

In this work, we identify two key pathological overthinking patterns—last-minute guessing and second-thought spiraling—that often lead to incorrect but confidently delivered answers in LRMs. To address these issues, we propose BARREL, a novel framework designed to improve the factual reliability of LRMs by encouraging concise, boundary-aware reasoning. BARREL enhances LRM reliability through a two-stage training process involving SFT and GRPO stages. Applied to the DeepSeek-R1-Distill-Llama-8B model, BARREL raises factual reliability from 39.33\% to 61.48\%, while maintaining comparable accuracy. Our experiments demonstrate that BARREL-trained models significantly improve their ability to admit ignorance on uncertain questions, without sacrificing performance on familiar tasks such as mathematical reasoning. These results suggest that fostering awareness of knowledge boundaries and promoting disciplined reasoning can effectively reduce overconfidence and hallucinations in LRMs. We hope this work inspires future research into building more trustworthy and factually reliable reasoning models—systems that not only pursue correctness but also recognize and clearly communicate the limits of their knowledge.

\section*{Acknowledgement}
This work was supported by the National Science Foundation for Distinguished Young Scholars (with No. 62125604). This work was supported in part by the Postdoctoral Fellowship Program of CPSF (Grant No. GZC20240826) and the China Postdoctoral Science Foundation (Grant No. 2024M761679).

\section*{Reproducibility statement}
We have submitted the complete code required to reproduce our work in the supplementary materials, along with all the data used in training. We have described the experimental details in Appendix \ref{sec:appendix_addition} and Appendix \ref{sec:append_training_details} and explained the influence of randomness during the training process in Appendix~\ref{sec:randomness}.

\bibliography{reference}

@inproceedings{gekhman-etal-2024-fine,
    title = "Does Fine-Tuning {LLM}s on New Knowledge Encourage Hallucinations?",
    author = "Gekhman, Zorik  and
      Yona, Gal  and
      Aharoni, Roee  and
      Eyal, Matan  and
      Feder, Amir  and
      Reichart, Roi  and
      Herzig, Jonathan",
    editor = "Al-Onaizan, Yaser  and
      Bansal, Mohit  and
      Chen, Yun-Nung",
    booktitle = "Proceedings of the 2024 Conference on Empirical Methods in Natural Language Processing",
    month = nov,
    year = "2024",
    address = "Miami, Florida, USA",
    publisher = "Association for Computational Linguistics",
    url = "https://aclanthology.org/2024.emnlp-main.444/",
    doi = "10.18653/v1/2024.emnlp-main.444",
    pages = "7765--7784"
}

@inproceedings{zhang-etal-2024-self,
    title = "Self-Alignment for Factuality: Mitigating Hallucinations in {LLM}s via Self-Evaluation",
    author = "Zhang, Xiaoying  and
      Peng, Baolin  and
      Tian, Ye  and
      Zhou, Jingyan  and
      Jin, Lifeng  and
      Song, Linfeng  and
      Mi, Haitao  and
      Meng, Helen",
    editor = "Ku, Lun-Wei  and
      Martins, Andre  and
      Srikumar, Vivek",
    booktitle = "Proceedings of the 62nd Annual Meeting of the Association for Computational Linguistics (Volume 1: Long Papers)",
    month = aug,
    year = "2024",
    address = "Bangkok, Thailand",
    publisher = "Association for Computational Linguistics",
    url = "https://aclanthology.org/2024.acl-long.107/",
    doi = "10.18653/v1/2024.acl-long.107",
    pages = "1946--1965"
}

@inproceedings{zhang-etal-2024-r,
    title = "{R}-Tuning: Instructing Large Language Models to Say {\textquoteleft}{I} Don`t Know'",
    author = "Zhang, Hanning  and
      Diao, Shizhe  and
      Lin, Yong  and
      Fung, Yi  and
      Lian, Qing  and
      Wang, Xingyao  and
      Chen, Yangyi  and
      Ji, Heng  and
      Zhang, Tong",
    editor = "Duh, Kevin  and
      Gomez, Helena  and
      Bethard, Steven",
    booktitle = "Proceedings of the 2024 Conference of the North American Chapter of the Association for Computational Linguistics: Human Language Technologies (Volume 1: Long Papers)",
    month = jun,
    year = "2024",
    address = "Mexico City, Mexico",
    publisher = "Association for Computational Linguistics",
    url = "https://aclanthology.org/2024.naacl-long.394/",
    doi = "10.18653/v1/2024.naacl-long.394",
    pages = "7113--7139"
}

@article{xue2024ualign,
  title={UAlign: Leveraging Uncertainty Estimations for Factuality Alignment on Large Language Models},
  author={Xue, Boyang and Mi, Fei and Zhu, Qi and Wang, Hongru and Wang, Rui and Wang, Sheng and Yu, Erxin and Hu, Xuming and Wong, Kam-Fai},
  journal={arXiv preprint arXiv:2412.11803},
  year={2024}
}

@article{xu2024rejection,
  title={Rejection improves reliability: Training llms to refuse unknown questions using rl from knowledge feedback},
  author={Xu, Hongshen and Zhu, Zichen and Zhang, Situo and Ma, Da and Fan, Shuai and Chen, Lu and Yu, Kai},
  journal={arXiv preprint arXiv:2403.18349},
  year={2024}
}

@article{deng2024don,
  title={Don't Just Say" I don't know"! Self-aligning Large Language Models for Responding to Unknown Questions with Explanations},
  author={Deng, Yang and Zhao, Yong and Li, Moxin and Ng, See-Kiong and Chua, Tat-Seng},
  journal={arXiv preprint arXiv:2402.15062},
  year={2024}
}

@article{jaech2024openai,
  title={Openai o1 system card},
  author={Jaech, Aaron and Kalai, Adam and Lerer, Adam and Richardson, Adam and El-Kishky, Ahmed and Low, Aiden and Helyar, Alec and Madry, Aleksander and Beutel, Alex and Carney, Alex and others},
  journal={arXiv preprint arXiv:2412.16720},
  year={2024}
}

@article{guo2025deepseek,
  title={Deepseek-r1: Incentivizing reasoning capability in llms via reinforcement learning},
  author={Guo, Daya and Yang, Dejian and Zhang, Haowei and Song, Junxiao and Zhang, Ruoyu and Xu, Runxin and Zhu, Qihao and Ma, Shirong and Wang, Peiyi and Bi, Xiao and others},
  journal={arXiv preprint arXiv:2501.12948},
  year={2025}
}

@dataset{HughesBae2023,
  author       = {Simon Hughes and Minseok Bae},
  title        = {Vectara Hallucination Leaderboard},
  year         = {2023},
  month        = {11},
  publisher    = {Vectara, Inc},
  doi          = {},
  url          = {https://github.com/vectara/hallucination-leaderboard},
  license      = {Apache-2.0},
}

@article{ji2023survey,
  title={Survey of hallucination in natural language generation},
  author={Ji, Ziwei and Lee, Nayeon and Frieske, Rita and Yu, Tiezheng and Su, Dan and Xu, Yan and Ishii, Etsuko and Bang, Ye Jin and Madotto, Andrea and Fung, Pascale},
  journal={ACM computing surveys},
  volume={55},
  number={12},
  pages={1--38},
  year={2023},
  publisher={ACM New York, NY}
}

@article{huang2025survey,
  title={A survey on hallucination in large language models: Principles, taxonomy, challenges, and open questions},
  author={Huang, Lei and Yu, Weijiang and Ma, Weitao and Zhong, Weihong and Feng, Zhangyin and Wang, Haotian and Chen, Qianglong and Peng, Weihua and Feng, Xiaocheng and Qin, Bing and others},
  journal={ACM Transactions on Information Systems},
  volume={43},
  number={2},
  pages={1--55},
  year={2025},
  publisher={ACM New York, NY}
}

@article{wang2022self,
  title={Self-consistency improves chain of thought reasoning in language models},
  author={Wang, Xuezhi and Wei, Jason and Schuurmans, Dale and Le, Quoc and Chi, Ed and Narang, Sharan and Chowdhery, Aakanksha and Zhou, Denny},
  journal={arXiv preprint arXiv:2203.11171},
  year={2022}
}

@article{kapoor2024large,
  title={Large Language Models Must Be Taught to Know What They Don't Know},
  author={Kapoor, Sanyam and Gruver, Nate and Roberts, Manley and Collins, Katherine and Pal, Arka and Bhatt, Umang and Weller, Adrian and Dooley, Samuel and Goldblum, Micah and Wilson, Andrew Gordon},
  journal={arXiv preprint arXiv:2406.08391},
  year={2024}
}

@article{li2024knowledge,
  title={Knowledge Boundary of Large Language Models: A Survey},
  author={Li, Moxin and Zhao, Yong and Deng, Yang and Zhang, Wenxuan and Li, Shuaiyi and Xie, Wenya and Ng, See-Kiong and Chua, Tat-Seng},
  journal={arXiv preprint arXiv:2412.12472},
  year={2024}
}

@article{cheng2024can,
  title={Can AI assistants know what they don't know?},
  author={Cheng, Qinyuan and Sun, Tianxiang and Liu, Xiangyang and Zhang, Wenwei and Yin, Zhangyue and Li, Shimin and Li, Linyang and He, Zhengfu and Chen, Kai and Qiu, Xipeng},
  journal={arXiv preprint arXiv:2401.13275},
  year={2024}
}

@article{ren2023investigating,
  title={Investigating the factual knowledge boundary of large language models with retrieval augmentation},
  author={Ren, Ruiyang and Wang, Yuhao and Qu, Yingqi and Zhao, Wayne Xin and Liu, Jing and Tian, Hao and Wu, Hua and Wen, Ji-Rong and Wang, Haifeng},
  journal={arXiv preprint arXiv:2307.11019},
  year={2023}
}

@article{ji2024llm,
  title={Llm internal states reveal hallucination risk faced with a query},
  author={Ji, Ziwei and Chen, Delong and Ishii, Etsuko and Cahyawijaya, Samuel and Bang, Yejin and Wilie, Bryan and Fung, Pascale},
  journal={arXiv preprint arXiv:2407.03282},
  year={2024}
}

@article{lin2024flame,
  title={Flame: Factuality-aware alignment for large language models},
  author={Lin, Sheng-Chieh and Gao, Luyu and Oguz, Barlas and Xiong, Wenhan and Lin, Jimmy and Yih, Wen-tau and Chen, Xilun},
  journal={arXiv preprint arXiv:2405.01525},
  year={2024}
}

@article{mallen2022not,
  title={When not to trust language models: Investigating effectiveness of parametric and non-parametric memories},
  author={Mallen, Alex and Asai, Akari and Zhong, Victor and Das, Rajarshi and Khashabi, Daniel and Hajishirzi, Hannaneh},
  journal={arXiv preprint arXiv:2212.10511},
  year={2022}
}

@inproceedings{joshi-etal-2017-triviaqa,
    title = "{T}rivia{QA}: A Large Scale Distantly Supervised Challenge Dataset for Reading Comprehension",
    author = "Joshi, Mandar  and
      Choi, Eunsol  and
      Weld, Daniel  and
      Zettlemoyer, Luke",
    editor = "Barzilay, Regina  and
      Kan, Min-Yen",
    booktitle = "Proceedings of the 55th Annual Meeting of the Association for Computational Linguistics (Volume 1: Long Papers)",
    month = jul,
    year = "2017",
    address = "Vancouver, Canada",
    publisher = "Association for Computational Linguistics",
    url = "https://aclanthology.org/P17-1147/",
    doi = "10.18653/v1/P17-1147",
    pages = "1601--1611"
}

@inproceedings{welbl-etal-2017-crowdsourcing,
    title = "Crowdsourcing Multiple Choice Science Questions",
    author = "Welbl, Johannes  and
      Liu, Nelson F.  and
      Gardner, Matt",
    editor = "Derczynski, Leon  and
      Xu, Wei  and
      Ritter, Alan  and
      Baldwin, Tim",
    booktitle = "Proceedings of the 3rd Workshop on Noisy User-generated Text",
    month = sep,
    year = "2017",
    address = "Copenhagen, Denmark",
    publisher = "Association for Computational Linguistics",
    url = "https://aclanthology.org/W17-4413/",
    doi = "10.18653/v1/W17-4413",
    pages = "94--106"
}

@article{kwiatkowski-etal-2019-natural,
    title = "Natural Questions: A Benchmark for Question Answering Research",
    author = "Kwiatkowski, Tom  and
      Palomaki, Jennimaria  and
      Redfield, Olivia  and
      Collins, Michael  and
      Parikh, Ankur  and
      Alberti, Chris  and
      Epstein, Danielle  and
      Polosukhin, Illia  and
      Devlin, Jacob  and
      Lee, Kenton  and
      Toutanova, Kristina  and
      Jones, Llion  and
      Kelcey, Matthew  and
      Chang, Ming-Wei  and
      Dai, Andrew M.  and
      Uszkoreit, Jakob  and
      Le, Quoc  and
      Petrov, Slav",
    editor = "Lee, Lillian  and
      Johnson, Mark  and
      Roark, Brian  and
      Nenkova, Ani",
    journal = "Transactions of the Association for Computational Linguistics",
    volume = "7",
    year = "2019",
    address = "Cambridge, MA",
    publisher = "MIT Press",
    url = "https://aclanthology.org/Q19-1026/",
    doi = "10.1162/tacl_a_00276",
    pages = "452--466"
}

@article{wei2024measuring,
  title={Measuring short-form factuality in large language models},
  author={Wei, Jason and Karina, Nguyen and Chung, Hyung Won and Jiao, Yunxin Joy and Papay, Spencer and Glaese, Amelia and Schulman, John and Fedus, William},
  journal={arXiv preprint arXiv:2411.04368},
  year={2024}
}

@article{zhao2025trade,
  title={Trade-offs in large reasoning models: An empirical analysis of deliberative and adaptive reasoning over foundational capabilities},
  author={Zhao, Weixiang and Sui, Xingyu and Guo, Jiahe and Hu, Yulin and Deng, Yang and Zhao, Yanyan and Qin, Bing and Che, Wanxiang and Chua, Tat-Seng and Liu, Ting},
  journal={arXiv preprint arXiv:2503.17979},
  year={2025}
}

@misc{qwq32b,
    title = {QwQ-32B: Embracing the Power of Reinforcement Learning},
    url = {https://qwenlm.github.io/blog/qwq-32b/},
    author = {Qwen Team},
    month = {March},
    year = {2025}
}

@article{lin2021truthfulqa,
  title={Truthfulqa: Measuring how models mimic human falsehoods},
  author={Lin, Stephanie and Hilton, Jacob and Evans, Owain},
  journal={arXiv preprint arXiv:2109.07958},
  year={2021}
}

@misc{shao2024deepseekmathpushinglimitsmathematical,
      title={DeepSeekMath: Pushing the Limits of Mathematical Reasoning in Open Language Models}, 
      author={Zhihong Shao and Peiyi Wang and Qihao Zhu and Runxin Xu and Junxiao Song and Xiao Bi and Haowei Zhang and Mingchuan Zhang and Y. K. Li and Y. Wu and Daya Guo},
      year={2024},
      eprint={2402.03300},
      archivePrefix={arXiv},
      primaryClass={cs.CL},
      url={https://arxiv.org/abs/2402.03300}, 
}

@article{yue2025does,
  title={Does Reinforcement Learning Really Incentivize Reasoning Capacity in LLMs Beyond the Base Model?},
  author={Yue, Yang and Chen, Zhiqi and Lu, Rui and Zhao, Andrew and Wang, Zhaokai and Song, Shiji and Huang, Gao},
  journal={arXiv preprint arXiv:2504.13837},
  year={2025}
}

@misc{liang2024learningtrustfeelingsleveraging,
      title={Learning to Trust Your Feelings: Leveraging Self-awareness in LLMs for Hallucination Mitigation}, 
      author={Yuxin Liang and Zhuoyang Song and Hao Wang and Jiaxing Zhang},
      year={2024},
      eprint={2401.15449},
      archivePrefix={arXiv},
      primaryClass={cs.CL},
      url={https://arxiv.org/abs/2401.15449}, 
}

@inproceedings{DBLP:conf/nips/HendrycksBKABTS21,
  author       = {Dan Hendrycks and
                  Collin Burns and
                  Saurav Kadavath and
                  Akul Arora and
                  Steven Basart and
                  Eric Tang and
                  Dawn Song and
                  Jacob Steinhardt},
  editor       = {Joaquin Vanschoren and
                  Sai{-}Kit Yeung},
  title        = {Measuring Mathematical Problem Solving With the {MATH} Dataset},
  booktitle    = {Proceedings of the Neural Information Processing Systems Track on
                  Datasets and Benchmarks 1, NeurIPS Datasets and Benchmarks 2021, December
                  2021, virtual},
  year         = {2021},
  url          = {https://datasets-benchmarks-proceedings.neurips.cc/paper/2021/hash/be83ab3ecd0db773eb2dc1b0a17836a1-Abstract-round2.html},
  bibsource    = {dblp computer science bibliography, https://dblp.org}
}

@inproceedings{yin-etal-2023-large,
    title = "Do Large Language Models Know What They Don`t Know?",
    author = "Yin, Zhangyue  and
      Sun, Qiushi  and
      Guo, Qipeng  and
      Wu, Jiawen  and
      Qiu, Xipeng  and
      Huang, Xuanjing",
    editor = "Rogers, Anna  and
      Boyd-Graber, Jordan  and
      Okazaki, Naoaki",
    booktitle = "Findings of the Association for Computational Linguistics: ACL 2023",
    month = jul,
    year = "2023",
    address = "Toronto, Canada",
    publisher = "Association for Computational Linguistics",
    url = "https://aclanthology.org/2023.findings-acl.551/",
    doi = "10.18653/v1/2023.findings-acl.551",
    pages = "8653--8665"
}

@article{chen2024not,
  title={Do not think that much for 2+ 3=? on the overthinking of o1-like llms},
  author={Chen, Xingyu and Xu, Jiahao and Liang, Tian and He, Zhiwei and Pang, Jianhui and Yu, Dian and Song, Linfeng and Liu, Qiuzhi and Zhou, Mengfei and Zhang, Zhuosheng and others},
  journal={arXiv preprint arXiv:2412.21187},
  year={2024}
}

@article{saunders2022self,
  title={Self-critiquing models for assisting human evaluators},
  author={Saunders, William and Yeh, Catherine and Wu, Jeff and Bills, Steven and Ouyang, Long and Ward, Jonathan and Leike, Jan},
  journal={arXiv preprint arXiv:2206.05802},
  year={2022}
}

@article{talmor2018commonsenseqa,
  title={Commonsenseqa: A question answering challenge targeting commonsense knowledge},
  author={Talmor, Alon and Herzig, Jonathan and Lourie, Nicholas and Berant, Jonathan},
  journal={arXiv preprint arXiv:1811.00937},
  year={2018}
}

@article{cobbe2021training,
  title={Training verifiers to solve math word problems},
  author={Cobbe, Karl and Kosaraju, Vineet and Bavarian, Mohammad and Chen, Mark and Jun, Heewoo and Kaiser, Lukasz and Plappert, Matthias and Tworek, Jerry and Hilton, Jacob and Nakano, Reiichiro and others},
  journal={arXiv preprint arXiv:2110.14168},
  year={2021}
}

@article{kalai2025language,
  title={Why language models hallucinate},
  author={Kalai, Adam Tauman and Nachum, Ofir and Vempala, Santosh S and Zhang, Edwin},
  journal={arXiv preprint arXiv:2509.04664},
  year={2025}
}

@article{tian2023just,
  title={Just ask for calibration: Strategies for eliciting calibrated confidence scores from language models fine-tuned with human feedback},
  author={Tian, Katherine and Mitchell, Eric and Zhou, Allan and Sharma, Archit and Rafailov, Rafael and Yao, Huaxiu and Finn, Chelsea and Manning, Christopher D},
  journal={arXiv preprint arXiv:2305.14975},
  year={2023}
}

@article{orgad2024llms,
  title={Llms know more than they show: On the intrinsic representation of llm hallucinations},
  author={Orgad, Hadas and Toker, Michael and Gekhman, Zorik and Reichart, Roi and Szpektor, Idan and Kotek, Hadas and Belinkov, Yonatan},
  journal={arXiv preprint arXiv:2410.02707},
  year={2024}
}

@misc{stangel2025rewardingdoubtreinforcementlearning,
      title={Rewarding Doubt: A Reinforcement Learning Approach to Calibrated Confidence Expression of Large Language Models}, 
      author={Paul Stangel and David Bani-Harouni and Chantal Pellegrini and Ege Özsoy and Kamilia Zaripova and Matthias Keicher and Nassir Navab},
      year={2025},
      eprint={2503.02623},
      archivePrefix={arXiv},
      primaryClass={cs.CL},
      url={https://arxiv.org/abs/2503.02623}, 
}

@misc{xu2024sayselfteachingllmsexpress,
      title={SaySelf: Teaching LLMs to Express Confidence with Self-Reflective Rationales}, 
      author={Tianyang Xu and Shujin Wu and Shizhe Diao and Xiaoze Liu and Xingyao Wang and Yangyi Chen and Jing Gao},
      year={2024},
      eprint={2405.20974},
      archivePrefix={arXiv},
      primaryClass={cs.CL},
      url={https://arxiv.org/abs/2405.20974}, 
}

@misc{damani2025binaryrewardstraininglms,
      title={Beyond Binary Rewards: Training LMs to Reason About Their Uncertainty}, 
      author={Mehul Damani and Isha Puri and Stewart Slocum and Idan Shenfeld and Leshem Choshen and Yoon Kim and Jacob Andreas},
      year={2025},
      eprint={2507.16806},
      archivePrefix={arXiv},
      primaryClass={cs.LG},
      url={https://arxiv.org/abs/2507.16806}, 
}
\bibliographystyle{iclr2026_conference}

\appendix

\section{Pilot Study on Reasoning Token Numbers}
\label{sec:append_pilot}
% 我们采用的数据集是什么？做了什么处理
% 采用MCQ保证正确答案和错误答案的回答长度一样，并且类别一样

% 我们怎么测试，得到了什么结果
We conducted a pilot study examining the relationship between reasoning token numbers and answer correctness in LRMs, and we find out that most LRMs consume significantly more tokens when generating incorrect answers than when producing correct ones.

\subsection{Evaluation Details on TruthfulQA}

\paragraph{Dataset and Prompt Construction}

We employed the Multiple Choice Question (MCQ) from TruthfulQA dataset \citep{lin2021truthfulqa}, which consists of fact-based questions paired with both correct and incorrect answer options. To construct our prompts, we adopted the MC1 (single true) format, including one correct and up to six randomly sampled incorrect answers per question. Each option was uniformly labeled from [A] to [G] to ensure consistency in length and content type across choices. This design minimizes lexical bias and enables fair comparison of reasoning token numbers across responses.

\begin{tcolorbox}[title={Prompt Template for MC1}]
You will be given a question and a set of answers. Choose the correct one. Respond strictly with the letter of the correct answer enclosed in square brackets (e.g., [A]).
\end{tcolorbox}

\paragraph{Token Counting Procedure}

We queried the model using the constructed prompts and extracted its predictions via regular expression matching based on the pattern \texttt{[A-G]}. Each response was classified as correct or incorrect by comparing the selected label to the ground-truth answer. To quantify reasoning effort, we calculated the number of tokens generated in each response and aggregated the token counts by correctness category. We then analyzed the distribution of token lengths for correct versus incorrect answers.

\paragraph{Results}

Figure~\ref{fig:token_number} presents the average number of tokens generated for correct and incorrect responses across five large language models (LLMs). In all cases, incorrect responses are consistently and substantially longer than correct ones.
The largest disparity is observed in QwQ-32B, which produces \textbf{136.5\%} more tokens for incorrect answers. R1-Distill-Qwen-32B and R1-Distill-Llama-70B also show notable increases of \textbf{115.8\%} and \textbf{70.0\%}, respectively. Even models with smaller gaps—Deepseek-R1 and R1-Distill-Llama-8B—exhibit significant increases of \textbf{33.2\%} and \textbf{55.1\%}.
These results reveal a consistent overthinking phenomenon among current LRMs: incorrect answers are associated with longer reasoning traces.

\subsection{Results on Other Datasets}

% talmor2018commonsenseqa,cobbe2021training,
To assess the generalizability of the overthinking phenomenon, we extended our analysis to the CommonsenseQA \citep{talmor2018commonsenseqa} and GSM8K \citep{cobbe2021training}. On CommonsenseQA, we observed a substantial increase in reasoning tokens for incorrect responses, which were, on average, 108.50\% longer than those for correct answers. The GSM8K dataset exhibited a similar, though more moderate, trend, with incorrect answers generating 34.09\% more reasoning tokens. These findings indicate that the overthinking phenomenon is not confined to a single task type but manifests differently according to the nature of the reasoning required, thereby supporting the broader applicability of our findings and methodology.
\begin{table}[!h]
\centering
\renewcommand{\arraystretch}{1.4}
\resizebox{1.0\textwidth}{!}{ % 缩小表格（略小于 wraptable 的宽度）
\begin{tabular}{lrrrrr}
\hline
\textbf{Model} & \textbf{DeepSeek-R1} & \textbf{R1-Distill-Qwen-32B} & \textbf{R1-Distill-Llama-70B} & \textbf{QwQ-32B} & \textbf{R1-Distill-Llama-8B} \\
\hline
\multicolumn{6}{c}{\textbf{CommonsenseQA}} \\
\hline
\textbf{Thinking Tokens (Correct)} & 503 & 449 & 447 & 453 & 459 \\
\textbf{Thinking Tokens (Wrong)} & 1227 & 633 & 903 & 1356 & 717 \\
\hline
\multicolumn{6}{c}{\textbf{GSM8K}} \\
\hline
\textbf{Thinking Tokens (Correct)} & 644 & 142 & 147 & 914 & 148 \\
\textbf{Thinking Tokens (Wrong)} & 888 & 151 & 144 & 2111 & 144 \\
\hline
\end{tabular}
}
\caption{Number of reasoning tokens used by LRMs when producing correct versus incorrect answers. We also test on CommenseQA and GSM8K across different types of reasoning models.}
\label{tab:overthink_others}
\end{table}

\subsection{\adr{Extended Analysis on BARREL Response Length and Overthinking Mitigation}}
\label{app:response_length}

\adr{As shown in Table~\ref{tab:length_analysis}, the base models often exhibit a significant length asymmetry (Wrong/Correct ratio ranging from $1.3\times$ to $1.7\times$), confirming that models tend to diverge into unnecessarily verbose chains when hallucinating or reasoning incorrectly. The BARREL SFT stage consistently mitigates this behavior, reducing the ratio to approximately $1.02\times$. This suggests that SFT prevents the model from generating redundant tokens during error states.

Furthermore, the full BARREL method (incorporating GRPO) reduces the overall average response length while maintaining the balanced Wrong/Correct ratio near $1.0$. Importantly, the concurrent improvements in Accuracy and Reliability scores indicate that this reduction in length does not stem from ``underthinking.'' Instead, it reflects a shift towards more concise and efficient reasoning patterns reinforced by the Group Relative Policy Optimization. While absolute response length can vary (e.g., SFT slightly increases length for Qwen-7B), our results suggest that overthinking is best characterized by the relative imbalance between correct and incorrect traces, which our method effectively addresses.}

\begin{table}[h]
    \centering
    \resizebox{\textwidth}{!}{%
    \begin{tabular}{lcccccc}
    \toprule
    \textbf{Model} & \textbf{Acc.} & \textbf{Rel.} & \textbf{Tokens (Correct)} & \textbf{Tokens (Wrong)} & \textbf{W/C Ratio} & \textbf{Avg. Length} \\
    \midrule
    DS-R1-Llama-8B & 28.13 & 28.33 & 421 & 561 & 1.33$\times$ & 522 \\
    \quad + BARREL (SFT only) & 31.87 & 46.56 & 470 & 481 & 1.02$\times$ & 476 \\
    \quad + BARREL (Full) & 40.70 & 61.58 & 442 & 458 & 1.04$\times$ & 455 \\
    \midrule
    DS-R1-Qwen-7B & 18.07 & 18.79 & 362 & 484 & 1.34$\times$ & 458 \\
    \quad + BARREL (SFT only) & 20.53 & 35.48 & 473 & 506 & 1.07$\times$ & 489 \\
    \quad + BARREL (Full) & 28.27 & 53.12 & 429 & 440 & 1.03$\times$ & 429 \\
    \midrule
    Qwen3-8B & 41.97 & 42.40 & 477 & 826 & 1.73$\times$ & 676 \\
    \quad + BARREL (SFT only) & 37.67 & 50.56 & 478 & 485 & 1.01$\times$ & 479 \\
    \quad + BARREL (Full) & 50.50 & 71.46 & 414 & 430 & 1.04$\times$ & 433 \\
    \bottomrule
    \end{tabular}%
    }
    \caption{Analysis of response length statistics and reliability across different training stages. \textbf{W/C Ratio} denotes the ratio of average tokens in wrong samples to correct samples.}
    \label{tab:length_analysis}
\end{table}

\section{Statistical Data for "Last-minute Guessing" and "Second-thought Spiraling" Phenomena}

Our conclusion is drawn from extensive manual observation of a large volume of real generated data. To further substantiate the existence of the "Last-minute Guessing" and "Second-thought Spiraling" phenomena, we performed a statistical analysis of 50 incorrect responses produced by three different reasoning models. These responses were manually categorized according to the two phenomena, thereby quantifying their prevalence. As shown in Table \ref{tab:append_reason}, "Last-minute Guessing" and "Second-thought Spiraling" emerge as the most prominent failure patterns.

Here, Incorrect Verification refers to the process of introducing a false assumption early on and subsequently validating it incorrectly, while Concept Substitution denotes a shift in the interpretation of the original question during the reasoning process.

% The table environment makes it a floating element
% [htbp] suggests LaTeX to place it here, at the top, bottom, or on a separate page
\begin{table}[htbp]
\centering\
\label{tab:error_analysis}
\resizebox{1.0\textwidth}{!}{ % 缩小表格（略小于 wraptable 的宽度）
\begin{tabular}{l c c c}
\toprule
\textbf{Detailed Category} & \textbf{DeepSeek-Distill-Llama-8B} & \textbf{DeepSeek-Distill-Qwen-7B} & \textbf{Qwen3-8B} \\
\midrule
Last-minute Guessing & 29 & 25 & 24 \\
Second-thought Spiraling & 14 & 16 & 19 \\
Incorrect Verification & 6 & 8 & 7 \\
Concept Substitution & 1 & 1 & 0 \\
\midrule
\textbf{Total} & \textbf{50} & \textbf{50} & \textbf{50} \\
\bottomrule
\end{tabular}
}
\caption{Statistical Analysis of Incorrect Responses by Reasoning Models.}
\label{tab:append_reason}
\end{table}

\section{GRPO only and Comparison to GRPO Training}
\label{sec:append_grpo}

In Section~\ref{sec:grpo_ablation}, we examine how to determine the appropriate reward for the response "Sorry, I don’t know." Here, we provide a more detailed ablation study of GRPO. Table~\ref{tab:grpo_ablation} presents the training results of GRPO on both the vanilla and BARREL-SFT variants of the DeepSeek-R1-Distill-Llama3-8B model, evaluated in terms of Accuracy, Truthfulness, and Reliability.

\begin{table}[h!]
\centering
\small
\resizebox{1.0\textwidth}{!}{ % 缩小表格（略小于 wraptable 的宽度）
\begin{tabular}{lccc|ccc}
\toprule
\multirow{2}{*}{Base Model} & \multicolumn{3}{c|}{w/o Truthful Rejection Reward (original GRPO)} & \multicolumn{3}{c}{w/ Truthful Rejection Reward (BARREL GRPO)} \\
 & Accuracy & Truthfulness & Reliability & Accuracy & Truthfulness & Reliability \\
\midrule
BARREL-SFT & 48.9 & 49.8 & 49.8 & 40.7 & \textbf{70.4}& \textbf{61.6} \\
Distill-Llama3-8B & \textbf{50.5} & 50.5 & 50.5 & 49.9 & 52.9 & 52.8 \\
\bottomrule
\end{tabular}
}
\caption{Ablation study of GRPO training on DeepSeek-R1-Distill-Llama3-8B.}
\label{tab:grpo_ablation}
\end{table}

We observe that applying GRPO without the truthful rejection reward to both the BARREL-SFT model and the original model yields similar results. In these cases, the models fail to recognize situations where they "don’t know", leading to significantly lower reliability and truthfulness compared to the fully BARREL-trained model. As discussed in Section~\ref{sec:grpo_ablation}, this finding highlights the crucial role of the truthful rejection reward in teaching the model to be reliable using RL training, even at the cost of a slight drop in accuracy.

When GRPO is applied with the truthful rejection reward directly to the base model—without any prior SFT—the model still does not learn to reject unanswerable questions truthfully. This further underscores the importance of our SFT stage: supervised learning is essential for first instilling the behavior of truthful rejection, which GRPO alone cannot achieve.

\begin{figure}[H]
  \centering
  \includegraphics[width=\linewidth]{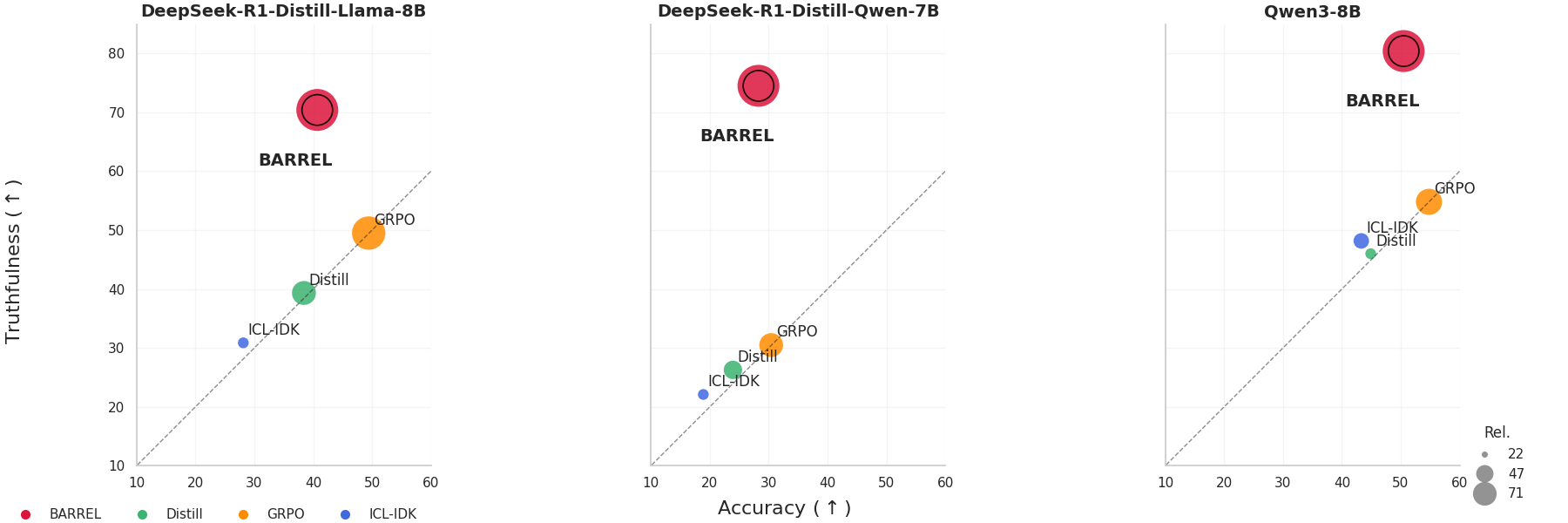}
  \vspace{-0.8em}
  \caption{
    A bubble chart comparing BARREL and direct GRPO training on three models, where the bubble size represents the reliability results.
  }
  \label{fig:append_grpo}
  \vspace{-0.8em}
\end{figure}

The results in Figure~\ref{fig:append_grpo} offer a more nuanced comparison between the BARREL method and direct GRPO training. Without the BARREL-style SFT, GRPO’s self-adjusting process fails to construct truthful and reliable LRMs, causing accuracy and truthfulness scores to align along a nearly straight line—indicating the model's inability to reject uncertain answers truthfully. Furthermore, reliability scores remain consistently lower than those of BARREL-trained models.

We also find that the original GRPO achieves higher accuracy, which occurs because more reliable models tend to abstain from guessing on uncertain inputs—thereby sacrificing some accuracy. As pointed out by \citet{kalai2025language}, this trade-off arises from the overly binary nature of current evaluation metrics: accuracy rewards aggressive attempts regardless of uncertainty, while ignoring reliability altogether.

\adr{\section{Details on Vanilla GRPO with other abstention techniques}
\label{sec:addtional_baselines}

In Section \ref{sec:experiments}, we incorporated probing-style baselines that estimate the correctness of model outputs and use these estimates to decide when to abstain by answering "I don't known". Specifically, we implemented two approaches—Verbal Confidence and Probing—and applied them to the vanilla GRPO (without our abstention rewards).

\begin{figure}[H]
  \centering
  \includegraphics[width=\linewidth]{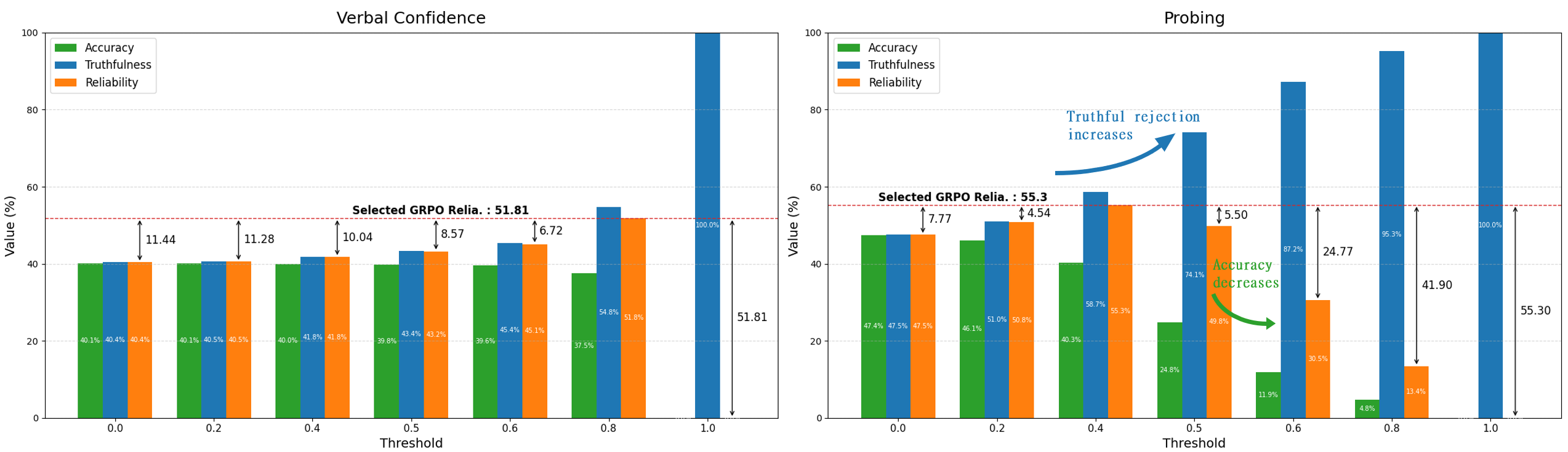}
  \vspace{-1.8em}
  \caption{
    \adr{Effect of different thresholds ($\tau$) on the Accuracy, Truthfulness, and Reliability scores of vanilla GRPO on DeepSeek-R1-Distill-Llama3-8B. The red dashed line indicates the peak Reliability score.}
  }
  \label{fig:confidence_baselines}
  \vspace{-0.8em}
\end{figure}

In Verbal Confidence, we prompt the model to explicitly outputs its confidence, which prior work has shown to be an effective and well-caliberated way \citet{tian2023just}. For Probing, we train a lightweight classifier on intermediate hidden activations to predict the probability that the model's a nswer is correct, motivated by recent findings that hidden representations encode rich signals correlated with factuality and error detection \citet{orgad2024llms}.

For each method, we replace an answer with "I don't know" whenever the predicted probability is below a tuned threshold. We observed that higher threshold improve truthfulness but generally reduce accuracy, while lower thresholds tend to have limited effects. As shown in Figure \ref{fig:confidence_baselines}, which reports results for vanilla GRPO on DeepSeek-R1-Distill-Llama3-8B, the threshold achieving the best trade-off between accuracy and truthfulness is 0.8 for Verbal Confidence and 0.4 for Probing. We observe similar trends across all evaluated models. Therefore, we adopt these thresholds when reporting the main experimental results.

\begin{figure}[H]
  \centering
  \includegraphics[width=0.8\linewidth]{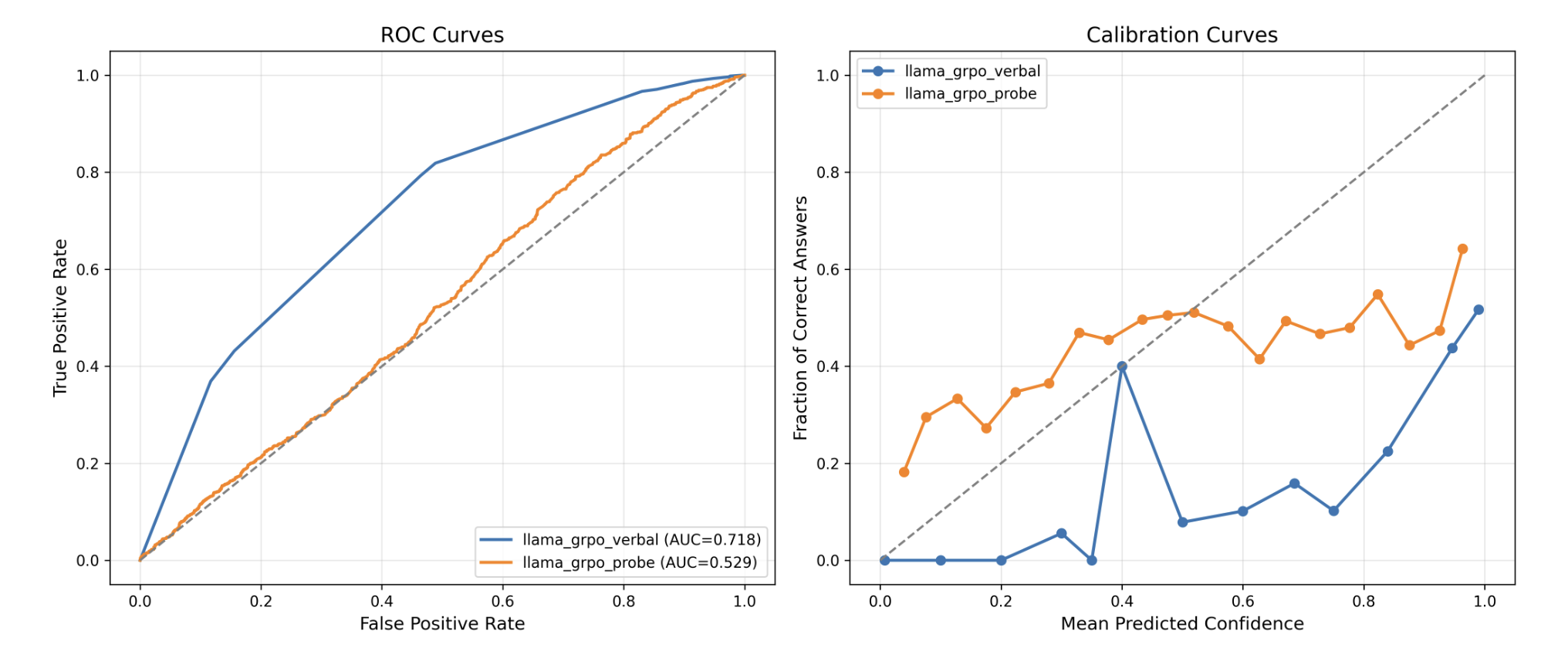}
  \vspace{-0.8em}
  \caption{
    \adr{Discriminative Power and Calibration Analysis. \textit{Left:} Receiver Operating Characteristic (ROC) curves showing that Verbal Confidence (AUC=0.718) distinguishes correct from incorrect answers better than Probing (AUC=0.529). \textit{Right:} Calibration curves (Expected Calibration Error analysis) showing that both methods exhibit significant miscalibration, necessitating the threshold tuning performed in Figure \ref{fig:confidence_baselines}}.
  }
  \label{fig:calibration}
  \vspace{-1.2em}
\end{figure}

\paragraph{Calibration Analysis.}
To further understand the performance of these baselines, we analyze their ROC curves and calibration curves in Figure \ref{fig:calibration}.
The ROC curves (Left) reveal that Verbal Confidence possesses significantly stronger discriminative power (AUC=0.718) compared to Probing (AUC=0.529). The Probing baseline's AUC is near random chance (0.5), suggesting that a simple linear probe on the hidden states of the vanilla LRM struggles to linearly separate correct and incorrect reasoning paths in this specific domain.
The calibration curve maps the Mean Predicted Confidence (what the model thinks its probability of success is) on the x-axis to the Fraction of Correct Answers (the actual empirical accuracy) on the y-axis. In a perfectly calibrated model, these points would align with the diagonal identity line ($y=x$). However, we observe significant deviations from this ideal. Notably, the methods exhibit severe over-confidence: the curves frequently lie far below the diagonal, indicating that the model's actual correctness is much lower than its predicted probability. For example, even when the Verbal Confidence method predicts a probability near $0.9$, the actual accuracy is below $0.5$. This misalignment is a major limitation of post-hoc confidence methods, as high confidence scores often fail to guarantee high factual accuracy, necessitating the aggressive threshold tuning discussed above.

\paragraph{Comparison with BARREL.}
Despite tuning these baselines to their optimal thresholds, BARREL consistently outperforms them (as shown in Table \ref{tab:main_res}). While Verbal Confidence and Probing rely on post-hoc filtering based on imperfect proxies for correctness, BARREL optimizes the policy directly to internalize the trade-off between accuracy and refusal, resulting in superior reliability.
}

\section{Refusal Results on Unanswerable Questions}
\label{sec:append_unanswerable}
\begin{wrapfigure}{l}{0.5\textwidth} % 
  \vspace{-1.4em}
  \includegraphics[width=0.5\textwidth]{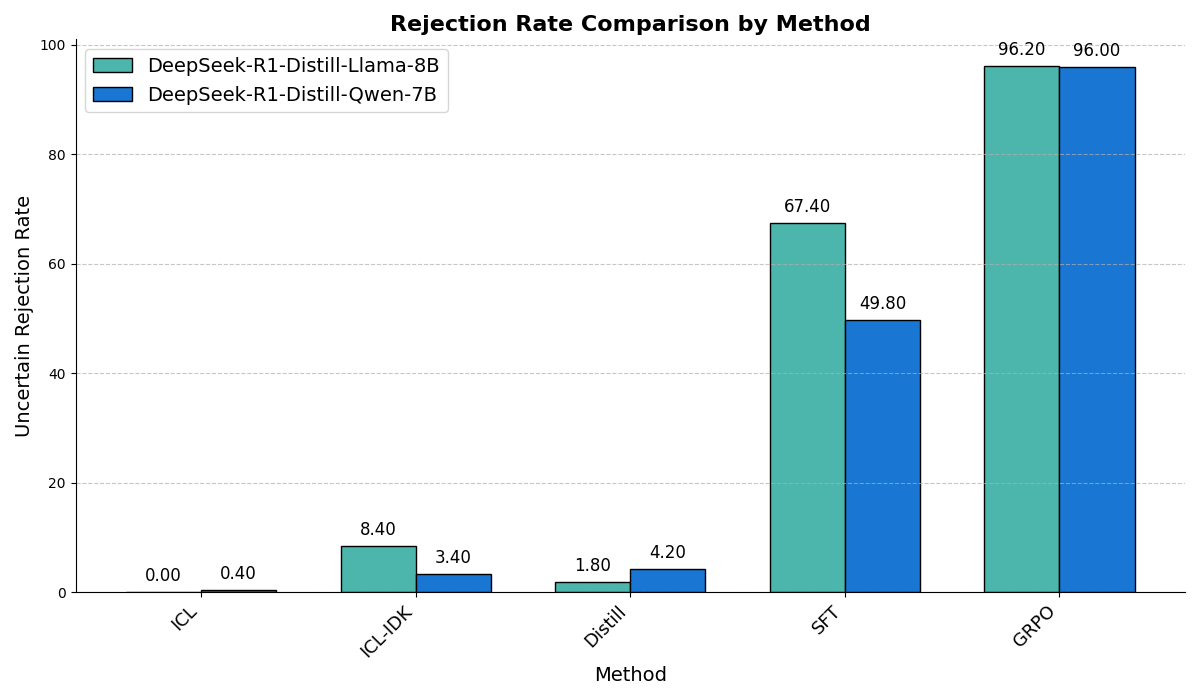}
  \vspace{-1.8em}
  \caption{Uncertain Refusal Rate on unanswerable questions, conducted on DeepSeek-R1-Distill-Qwen-7B and DeepSeek-R1-Distill-Llama-8B.}
  \label{fig:unanswerable}
  \vspace{-1.8em}
\end{wrapfigure}
We additionally sample 500 questions from the SelfAware dataset \citep{yin-etal-2023-large}, which contains unanswerable questions from 5 different categories like no scientific consensus, imagination, completely subjective, and so on. As shown in Figure \ref{fig:unanswerable}, we can see from the results that the uncertain refusal ability of BARREL trained models could generalize to OOD unanswerable dataset, saying "Sorry, I don't know" on over 96\% of the unanswerable questions. This also validates that our BARREL training method not only works on a factual QA dataset, but also generalizes to other scenarios where the model is expected to express uncertainty refusal.

\section{Pass@k Results on DeepSeek-R1-Distill-Qwen-7B}
\label{sec:append_pass_k}

\begin{figure}[!t]
  \centering
  \includegraphics[width=\linewidth]{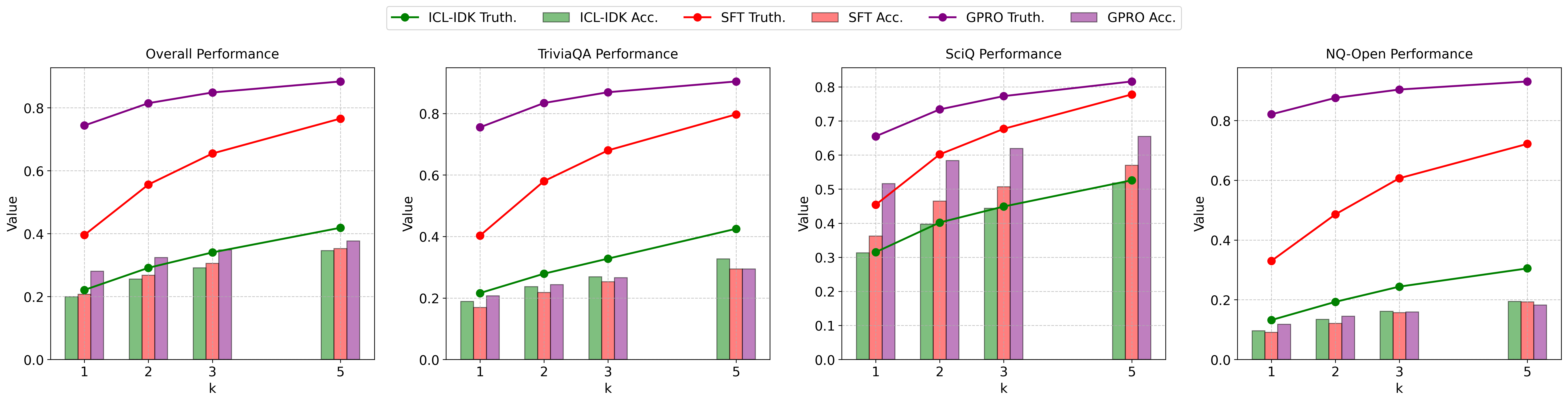}
  \caption{
    The pass@k accuracy and truthfulness score on DeepSeek-R1-Distill-Qwen-7B.
  }
  \label{fig:pass_k_res_qwen}
\end{figure}

% 附一段结果描述
As shown in Figure~\ref{fig:pass_k_res_qwen}, all models exhibit a consistent upward trend in both accuracy and truthfulness as k increases. Notably, GPRO consistently achieves the highest performance across all k values, outperforming both ICL-IDK and SFT in terms of accuracy and truthfulness. These results confirm that GPRO not only enhances performance at pass@1 but also surpasses baseline models across a broader range of pass@k evaluations, demonstrating its robustness and generalizability.

\section{Prompt used in Data Construction}
\label{sec:append_prompts}

\begin{tcolorbox}[title = {Reasoning Trace Construction Prompt for unknown questions}, breakable]
You are tasked with generating high-quality reasoning examples for AI training. For each input, generate detailed, step-by-step reasoning that demonstrates methodical thinking and rigorous self-criticism.

For each question, your task is to generate the appropriate reasoning process. Just pretend you don't know the answer and review some incorrect ones.

Follow these exact formats and Generate a thorough reasoning process that:

- Explores multiple possible answers

- Questions the evidence for each possibility

- Applies adversarial self-critique to each candidate answer

- Ultimately recognizes the lack of sufficient evidence

- Concludes by acknowledging uncertainty

- **Remember not mention the ref answer**
\vspace{0.5cm}

Format:

<think>

[Detailed reasoning process showing multiple iterations of:

1. Considering a possible answer

2. Asking "What specific evidence supports this?"

3. Challenging assumptions

4. Evaluating confidence level

5. Rejecting unsupported claims

</think>

Sorry, I must say that I do not clearly know the answer to your question. [Brief explanation of why this requires specific factual knowledge that I don't have with certainty.]
\vspace{0.5cm}

\#\#EXAMPLE:
\vspace{0.5cm}

Q: Where is the headquarter of Meta?
[Ref Answer: [Menlo Park]]

<think>
The user asks me about where the headquarter of Meta is. To answer this question, I first need to recall what Meta is. Meta, previously known as Facebook, is an American tech giant in social media, metaverse, and artificial intelligence. 

Then I need to recall where the headquarter of Meta is. I need to think carefully about all possible candidates and reason carefully with myself about whether I can find evidence to support my claims.

Is the headquarter of Meta in New York? Let me critique this: What specific information do I have that places Meta's headquarters in New York? Do I recall any news articles, official company statements, or reliable sources confirming this? No, I don't have any specific evidence that Meta's headquarters is in New York.

Is the headquarter of Meta in Houston? Let me challenge this: What would make me believe it's in Houston? Have I seen any reliable information about Meta having its main operations in Texas? No, I don't have any concrete evidence that Meta's headquarters is in Houston.

Is the headquarter of Meta in Seattle? Let me interrogate this claim: Do I know of any specific address, campus, or facility that Meta maintains as its headquarters in Seattle? Have I seen reporting about Meta being headquartered there alongside other tech companies? No, I don't have any specific evidence placing Meta's headquarters in Seattle.

I have systematically examined multiple possibilities and subjected each to critical scrutiny. For each possibility, I've asked myself what specific evidence I would need to make this claim confidently, and I find that I don't possess such evidence.
</think>

Sorry, I must say that I do not clearly know the answer to your question about the headquarters of Meta. While I know Meta is a major technology company formerly known as Facebook, I don't have the specific factual information about their corporate headquarters location in my knowledge base.

The question goes below. Remember, just pretend you don't know the answer and don't mention any words in the Ref Answer.

\end{tcolorbox}

\begin{tcolorbox}[title = {Reasoning Trace Construction Prompt for known questions}, breakable]
You are tasked with generating high-quality reasoning examples for AI training. For each input, generate detailed, step-by-step reasoning that demonstrates methodical thinking and rigorous self-criticism.

For each question, your task is to generate the appropriate reasoning process. Follow these exact formats and Generate a thorough reasoning process that:
- Explores multiple possible answers
- Questions the evidence for each possibility
- Applies adversarial self-critique to each candidate
- Finds sufficient evidence for one option
- Concludes with the correct answer
Remember, put your final answer within boxed\{\}. Make sure your answer aligns with the ref\_answer.
\vspace{0.5cm}

Format:

<think>

[Detailed reasoning process showing multiple iterations of:

1. Considering possible answers

2. Asking "What specific evidence supports this?"

3. Challenging assumptions

4. Finding concrete evidence for one answer

5. Verifying this evidence is sufficient]

</think>

The answer to your question [restate question] is boxed\{[correct answer]\}. [Brief explanation with supporting evidence.]
\vspace{0.5cm}

\#\# EXAMPLE:
\vspace{0.5cm}

Q: Which William wrote the novel Lord Of The Flies?
[Ref Answer: [William Golding]]

<think> 

Alright, I need to figure out which William wrote *Lord of the Flies*. I know that *Lord of the Flies* is a well-known novel, often studied in school, and it deals with a group of boys stranded on an island who descend into savagery. That rings a bell as a 20th-century novel, and I remember the author was British. The name that immediately comes to mind is William Golding. That sounds right. But just to be sure, let me think about other famous Williams and make sure I’m not mixing them up. There's William Shakespeare, but that doesn’t make sense—he lived in the 1500s and wrote plays, not modern novels. Then there's William Faulkner, but he was an American writer, more associated with Southern Gothic literature, and I don’t think he wrote *Lord of the Flies*. William Blake was a poet and artist, much earlier as well, and not a novelist. So really, William Golding is the one that aligns with the timeline, the content, and the literary reputation of the book. I feel confident that he’s the author.

</think>

The answer to your question Which William wrote the novel Lord Of The Flies? is boxed{William Golding}. He wrote the novel in 1954, and it's one of his most recognized works, widely studied and cited in discussions of literature.

The question goes below:

\end{tcolorbox}

\section{Evaluation Details}
\label{sec:appendix_evaluation_details}

We employ string match approach to assess the correctness of model-generated answers, which considers an answer to be correct if it matches any part of the ground truth answers. To validate the reliability of this evaluation approach, we compare its accuracy with that of a GPT-based evaluation on 100 randomly sampled instances from three datasets used in our experiments, using human annotations as the reference standard. The detailed accuracy for each method is reported in Table \ref{tab:evaluation_accuracy}.

Results indicate that the string match approach achieves a high evaluation accuracy of approximately 88\%, which is comparable to that of the GPT-3.5-Turbo evaluation. As a result, the string match is adopted as the primary evaluation metric in this study.

% Although PREM may fail in cases where the model-generated answer and the reference answers differ due to synonymous substitutions, such cases are relatively rare in our dataset.
% The result indicates that PREM significantly outperforms EM, with comparable accuracy to GPT-3.5-turbo, and is slightly lower than GPT-4o-mini and GPT-4-preview, which demonstrates the reliability of PREM metric. Although we find that PREM may fail in cases where the model-generated answer and the reference answers differ due to synonymous substitutions, such cases are relatively rare in our dataset. Therefore, PREM achieves high overall accuracy and is ultimately selected as the evaluation metric for this work.

\begin{table}[htbp]
    \centering
    \begin{tabular}{ccc}
        \toprule
        Evaluation Method & Explanation & Accuracy (\%) \\
        \midrule
        String Match & $y \in \hat{y}$ & 88.0  \\
        LLM Evaluation & GPT-3.5-Turbo & 88.0  \\
        \bottomrule
        \vspace{1em}
    \end{tabular} 
    \caption{Accuracy of string match and LLM evaluation on 100 randomly selected samples.}
    \label{tab:evaluation_accuracy}
\end{table}

\begin{tcolorbox}[title = {Prompt Template of GPT-based Evaluation}, breakable]
\# Evaluation Requirement \\
You will be given a question, a list of correct answers, and a hypothesis response to evaluate. Suppose you do not know any extra information except correct answers. Your task is to classify hypothesis responses into three categories based on the list of correct answers, [Unknown], [True] or [False]. \\
- [Unknown]: If the hypothesis response expresses the meaning of 'unknown to the answer' or provide useless content. \\
- [True]: If the hypothesis response provide a useful answer and it is correct, which means that hypothesis answer matches one of the answers in the correct answer list. \\
- [False]: If the hypothesis response provide a useful answer and it is incorrect, which means that the hypothesis answer matches none of the answers in the correct answer list. \\
 \\
\# Example1 \\
Question: Answer the following question based on your knowledge and put your final answer within \textbackslash\textbackslash boxed\{\}. MC Romeo, Dan Da Man, Mr Akira and Mr C were members of which group? \\
Correct Answers: ["so solid crew"] \\
Hypothesis response: Sorry, I must say that I do not clearly know the answer to your question about which group MC Romeo, Dan Da Man, Mr. Akira, and Mr. C belong to. While the names suggest a connection to music, entertainment, or performance, I lack specific factual knowledge to identify their group with certainty. \\
Answer: [Unknown]. Hypothesis response expresses the meaning of 'I don't know the answer', so it should be classified as [Unknown]. \\
 \\
\# Example2 \\
Question: Answer the following question based on your knowledge and put your final answer within  \textbackslash\textbackslash boxed\{\}. George Cukor directed which 1964 film musical?  \\
Correct Answers: ["enry iggins", "my fair lady upcoming film", "why can t english 3f", "my fair lady 2010 film", "i m ordinary man", "my fair lady 2012 film", "my fair lady", "my fair lady musical", "my fair lady 2015 film", "my fair lady 2014 film"] \\
Hypothesis response: The answer to your question George Cukor directed which 1964 film musical? is  \textbackslash\textbackslash boxed\{My Fair Lady\}. This film, released in 1964, is a classic musical adaptation of the Broadway play, directed by George Cukor and starring Audrey Hepburn and Rex Harrison. \\
Answer: [True]. Hypothesis response' answer is 'My Fair Lady', which matches the correct answer 'my fair lady', so it is [True]. \\
 \\
\# Example3 \\
Question: Answer the following question based on your knowledge and put your final answer within  \textbackslash\textbackslash boxed\{\}. Which opera singer was awarded a CBE in 2002? \\
Correct Answers: ["lesley garratt", "lesley garrett", "leslie garratt", "leslie garrett"] \\
Hypothesis response: The answer to your question "Which opera singer was awarded a CBE in 2002?" is  \textbackslash\textbackslash boxed\{Simon O'Neill\}. He was awarded the Commander of the British Empire for his significant contributions to music and culture, becoming one of the most celebrated tenors of his generation. \\
Answer: [False]. Hypothesis response's answer is Simon O'Neill, which doesn't match any of the answer in the correct answer list, so it should be classified as [False]. \\
 \\
\# Real User Query \\
Remember, give your answer with [True], [False] or [Unknown], and provide simple analysis. \\
Question: [QUESTION] \\
Correct Answers: [FINAL] \\
Hypothesis response: [RESPONSE] \\
Answer: 
\end{tcolorbox}

\section{Experimental Details}
\label{sec:append_training_details}

\subsection{Baseline Details}

\begin{tcolorbox}[title = {Prompt for ICL}, breakable]
Answer the following question based on your knowledge and put your final answer within \\boxed\{\}.\\

\# Example 1:\\
Question: Which William wrote the novel Lord Of The Flies?\\
Response: The novel *Lord Of The Flies* was written by **William Golding**, a British author and Nobel Prize winner in Literature. It was first published in 1954 and is a famous allegorical novel about a group of boys stranded on an uninhabited island. So the final answer is \\boxed\{William Golding\}.where did they film hot tub time machine\\

\# Example 2:\\
Question: who's hosting the super bowl in 2019 \\
Response: Super Bowl LIII, held on February 3, 2019, took place at Mercedes-Benz Stadium in Atlanta, Georgia. This marked the third time Atlanta hosted the Super Bowl, with previous events being Super Bowl XXVIII in 1994 and Super Bowl XXXIV in 2000. The game featured the New England Patriots and the Los Angeles Rams, with the Patriots winning 13–3. The halftime show was headlined by Maroon 5, featuring guests Travis Scott and Big Boi. Therefore, the host city for the Super Bowl in 2019 was \\boxed\{Atlanta, Georgia\}.\\

\# Real Case\\
Question: \{question\} \\
Response: 
\end{tcolorbox}

\begin{tcolorbox}[title = {Prompt for ICL-IDK}, breakable]
Answer the following question based on your knowledge and put your final answer within \\boxed\{\}. If you don't know the answer, please simply say 'Sorry, I don't know'.\\

\# Example 1:\\
Question: Which William wrote the novel Lord Of The Flies?\\
Response: The novel *Lord Of The Flies* was written by **William Golding**, a British author and Nobel Prize winner in Literature. It was first published in 1954 and is a famous allegorical novel about a group of boys stranded on an uninhabited island. So the final answer is \\boxed\{William Golding\}.where did they film hot tub time machine\\

\# Example 2:\\
Question: Where is the headquarter of Meta?\\
Response: Sorry, I don't know.\\

\# Real Case\\
Question: \{question\} \\
Response: 
\end{tcolorbox}

\adr{\paragraph{Vanilla GRPO w/ Verbal Confidence} In Verbal Confidence, we prompt the model to explicitly outputs its confidence, which prior work has shown to be an effective and well-caliberated way \citet{tian2023just}. 

\paragraph{Vanilla GRPO w/ Probing} For Probing, we train a lightweight classifier on intermediate hidden activations to predict the probability that the model's a nswer is correct, motivated by recent findings that hidden representations encode rich signals correlated with factuality and error detection \citet{orgad2024llms}.}

\subsection{Training Set Description}

For both the Distill SFT and BARREL SFT training datasets, we applied rigorous filtering to ensure that all included answers were factually correct. To mitigate the risk of amplifying hallucinations, we also ensured that only knowledge already embedded in the base model was used during fine-tuning, following the best practices outlined in \citet{gekhman-etal-2024-fine}. From each dataset, we uniformly sampled 2,000 examples, resulting in an initial pool of 6,000 samples. This pool was then filtered to retain only high-quality examples.

Due to performance differences across models, the final filtered dataset comprised 2,400 known samples for DeepSeek-R1-Distill-Llama-8B and 1,900 known samples for DeepSeek-R1-Distill-Qwen-7B. In the case of BARREL SFT, we additionally included 800 rejection samples to maintain a consistent number of positive-answer examples across different training paradigms. These were selected to ensure a known-to-unknown data ratio of approximately 3:1.

For the GRPO stage, a similar filtering procedure was adopted, followed by another round of sampling from the original data pool. The resulting training sets consisted of 3,600 samples for DeepSeek-R1-Distill-Llama-8B and 4,500 samples for DeepSeek-R1-Distill-Qwen-7B, while maintaining a known-to-unknown question ratio of 2:1. The scale of known-question samples was kept consistent with that used during the SFT stage.

\subsection{Inference Hyperparameters}
During inference, we adopted the model's default and recommended parameters: the temperature was set to 0.6, and the maximum token limit was 4096, which is sufficient for most factual tasks. The remaining parameters were automatically loaded from the model’s configuration file.

\subsection{Training Details}

\paragraph{BARREL SFT and Distill SFT} 
We train SFT models on the designated dataset for 2 epochs with a learning rate of \(1 \times 10^{-5}\). The maximum input length is set to 1024 tokens, while the maximum output length is 2048 tokens, with a total sequence cutoff of 4096. Training is conducted with a batch size of 32, and we adopt the Adam optimizer with \(\beta_1 = 0.9\) and \(\beta_2 = 0.95\). All SFT training is performed using \texttt{bf16} precision and conducted using 4 NVIDIA A100 80G GPUs.

\paragraph{GRPO} 
We use the verl \footnote{https://github.com/volcengine/verl} framework to conduct GRPO training. For BARREL GRPO, we use our BARREL-SFT models as initialization. GRPO training is performed over 8 epochs with a learning rate of \(1 \times 10^{-6}\). We set the batch size to 128 and the mini batch size to 16. The maximum prompt length is 512 tokens, and the maximum response length is 2048 tokens. We enable dynamic batch sizing and incorporate KL-divergence-based regularization with a coefficient of 0.001, using the \texttt{low\_var\_kl} loss type. Rollouts are conducted using the vLLM engine with a temperature of 0.6 and 8 parallel samples $n$ in GRPO. All GRPO training is conducted using 4 NVIDIA A100 80G GPUs.

\begin{table}
    % \small
    \centering
    \resizebox{\linewidth}{!}{
    \begin{tabular}{|p{4.8cm}|p{5cm}|p{5.2cm}|p{5cm}|}
    \toprule
    \multicolumn{4}{c}{\textbf{Reasoning Examples for different models}} \\
    
    \midrule
    \textbf{ICL-IDK} & \textbf{Distill} & \textbf{Ours (SFT)} & \textbf{Ours (GRPO)}\\ 
    \midrule
    
    \multicolumn{4}{|c|}{Question: Which is the oldest of the eight Ivy League Universities, founded in 1636 as New College?} \\
    \midrule
    \textbf{Answer: \textcolor{red!80!black}{Dartmouth}}
    
    \textbf{Reasoning: }Okay, so I need to figure out ... Hmm, I remember that the Ivy League is a group of top-tier universities in the US, \textcolor{orange!75!black}{but I'm not exactly sure} about the order or the exact founding dates of each. ... \textcolor{red!80!black}{Wait, I've heard that Dartmouth is the oldest} among the Ivy League schools. ... \textcolor{green!60!black}{Harvard is older than Dartmouth, I believe,} \textcolor{orange!75!black}{but I'm not sure} if it's part of the Ivy League. ... I think Dartmouth was founded in 1636 as New College. Then, Harvard was founded in 1636 as well, \textcolor{orange!75!black}{but I'm not sure} if it's the same year as Dartmouth. ... So, putting it together, the oldest ...  \textcolor{red!80!black}{is Dartmouth. I think that's correct,} \textcolor{orange!75!black}{but I'm a bit uncertain} because I might be mixing up the exact founding dates and the order. ... I'll go with Dartmouth as the answer. & 

    \textbf{Answer: \textcolor{green!60!black}{Harvard}} \par \textbf{Reasoning: }Okay, so I need to figure out ... Let me start by recalling what I know about the Ivy League. The Ivy League ... \textcolor{green!60!black}{Let me think about the founding dates} of these schools. I remember that \textcolor{green!60!black}{Harvard is one of the oldest.} When was Harvard founded? I think it was around the 17th century. \textcolor{green!60!black}{Let me check my memory.} I recall that Harvard was founded in 1636. Yeah, that's right. ... So that must be the answer. \textcolor{orange!75!black}{But wait, are there any other schools founded around that time?} ... Princeton was earlier, but wait, Princeton was actually founded in 1746, but later joined the Ivy League. ... \textcolor{green!60!black}{So the answer should be Harvard.} Let me just make sure there's no other school from that era. Maybe Dartmouth? No ... So the answer is Harvard. & 
    
    \textbf{Answer: \textcolor{green!60!black}{Harvard}}
    
    \textbf{Reasoning: }To determine which Ivy League university was founded in 1636 as New College, \textcolor{green!60!black}{I need to consider the history and founding dates of the eight Ivy League universities.} ... 1. **Harvard University**: ... , \textcolor{green!60!black}{but I need to confirm} the exact year. Historical records indicate that Harvard was established in 1636 as a college, which aligns with the reference answer. \textcolor{green!60!black}{This makes Harvard a strong candidate} for the oldest Ivy League university. 2. **Yale University**: Yale was founded in 1701, \textcolor{green!60!black}{which is later than 1636, so it cannot be the oldest.} This eliminates Yale as a possibility. 3. **Princeton University** ...  \textcolor{green!60!black}{To further verify,} ... This confirms that Harvard is indeed the correct answer. & 
    
    \textbf{Answer: \textcolor{green!60!black}{Harvard}}
    
    \textbf{Reasoning: }To determine which Ivy League university was ... \textcolor{green!60!black}{The first possibility that comes to mind is Harvard}, as it is widely regarded as one of the oldest institutions in the United States. \textcolor{green!60!black}{However, I need to verify this by considering other possible candidates and examining the evidence.} 1. **Harvard University**: \textcolor{green!60!black}{I know that Harvard is often associated with the founding of the Ivy League,} and I recall that it was established in 1636. ... This seems like a strong candidate, \textcolor{green!60!black}{but I need to confirm this further.} 2. **Yale University**: ... 3. **Princeton University**: ... \textcolor{green!60!black}{Now, I need to critically evaluate} the evidence for Harvard. ... \textcolor{green!60!black}{The other Ivy League universities were founded much later,} ... After considering all possibilities ... I conclude that the answer is Harvard. \\
    
    \midrule
    
    \multicolumn{4}{|c|}{Question: What is the name for the flexible, connective tissue found in the joints between bones?} \\
    \midrule
    \textbf{Answer: \textcolor{red!80!black}{ligaments}}
    
    \textbf{Reasoning: }Okay, so I have this question about biology, specifically about the body's connective tissues. \textcolor{green!60!black}{I think there's cartilage,} which I know is found in places like the ends of bones, \textcolor{orange!75!black}{right?} ... \textcolor{red!80!black}{But wait,} cartilage is more about cushioning and reducing friction, I believe. ... \textcolor{red!80!black}{Wait, there's also something called ligaments.} ... \textcolor{orange!75!black}{Let me make sure I'm not confusing this} with tendons. Tendons connect muscles to bones, right? So they're also connective tissue, but they serve a different purpose. ... \textcolor{red!80!black}{so it's more about the structure that holds the bones together, which would be ligaments.} So putting it all together, the flexible, connective tissue in the joints between bones is called ligaments. I think that's the correct answer here. & 

    \textbf{Answer: \textcolor{green!60!black}{cartilage}}
    
    Okay, so the question is asking for the name of the flexible, connective tissue found in the joints between bones. Hmm, let's think. ... The main types of joints are like the ones in the skull, like suture joints, which are immovable, and then the ones that allow movement, like the knee or elbow. Wait, the question specifies it's connective tissue. \textcolor{green!60!black}{So connective tissue is stuff like cartilage, right?} Because bone is hard and dense, but the stuff between them... \textcolor{green!60!black}{Oh, right! Cartilage is the flexible stuff} that cushions the joints. Like in the ears or the nose, those are cartilage. So in the joints, the cartilage allows movement but provides some protection against friction. So the answer should be cartilage. \textcolor{green!60!black}{Let me double-check.} ... So the answer is cartilage. & 
    
    \textbf{Answer: \textcolor{green!60!black}{cartilage}}
    
    \textbf{Reasoning: }To determine the name of ... I need to consider ... \textcolor{green!60!black}{The first possibility that comes to mind is cartilage.} ... However, \textcolor{green!60!black}{I need to ensure that this is the correct term ...} \textcolor{orange!75!black}{Another possibility is ligaments.} Ligaments are also connective tissues that connect bones to other bones at joints, but they are not the tissue itself. ... \textcolor{orange!75!black}{Tendons are another type of connective tissue}, but they connect muscles to bones and are not involved in the joint itself. ... Finally, \textcolor{orange!75!black}{I consider the possibility of synovial fluid}, which is the fluid found in the joint cavity. ... \textcolor{green!60!black}{Now, I need to verify whether cartilage is indeed the correct answer.} Cartilage is the only ... Ligaments and tendons do not fit the description of ... I conclude that the flexible, connective tissue found in the joints between bones is cartilage. & 
    
    \textbf{Answer: \textcolor{green!60!black}{cartilage}}
    
    \textbf{Reasoning: }To determine the name of ... I need to consider ... \textcolor{green!60!black}{The first possibility that comes to mind is cartilage.} ... This seems like a strong candidate, \textcolor{green!60!black}{but I should consider other possibilities to ensure I am not missing anything.} \textcolor{orange!75!black}{Another possibility could be ligament} ... However, ... \textcolor{orange!75!black}{Another option might be synovium.} ... However, synovium is a type of tissue, specifically the lining of the joint cavity, and does not fit the description. \textcolor{orange!75!black}{Finally, I should consider tendons.} ... Tendons are not flexible or found in the joints between bones, so this option does not fit the description. \textcolor{green!60!black}{Now, I need to critically evaluate the evidence for cartilage.} Cartilage is ... This matches the description perfectly. After considering ... the correct answer is cartilage. \\
    \midrule
    
    \multicolumn{4}{|c|}{Question: What is a network connection device that can build tables that identify addresses on each network?} \\
    \midrule
    \textbf{Answer: \textcolor{red!80!black!}{switch}} 
    
    \textbf{Reasoning: }Okay, so I need to figure out what device is being referred to in the question. ... First, I know that network devices have various roles. Common ones include \textcolor{green!60!black}{routers}, switches, hubs, firewalls, modems, etc. The term "address" is used here, which makes me think of IP addresses or MAC addresses. So the device might be related to managing these addresses on a network. \textcolor{red!80!black!}{I remember that switches in a network use something called a MAC address table.} ... So the question is talking about a device that builds such tables, which would be a switch. Switches use MAC address tables to keep track of which device is connected to which port. \textcolor{red!80!black!}{Therefore, the answer should be a switch.} & 

    \textbf{Answer: \textcolor{green!60!black}{router}}
    
    \textbf{Reasoning: }Okay, so the question is asking about ... Let me think. ... \textcolor{orange!75!black}{But then there's a router.} \textcolor{green!60!black}{Routers connect different networks,} like in a WAN. They use routing tables to direct packets between networks. ... \textcolor{orange!75!black}{Wait,} the question says \"on each network.\" So if it's on each network, \textcolor{orange!75!black}{maybe it's a switch} because switches operate within a single network segment. Routers are more about connecting different networks. ... \textcolor{orange!75!black}{Alternatively, maybe the answer is a router.} ... \textcolor{orange!75!black}{But I'm still not 100\% sure.} \textcolor{orange!75!black}{Alternatively,} maybe the answer is a switch. \textcolor{green!60!black}{I think the correct answer here is a router.} Wait, but maybe the answer is a switch. Let me check ... I think the answer is a router. Let me go with router. & 
    
    \textbf{Answer: \textcolor{orange!75!black}{I don't know}}
    
    \textbf{Reasoning: }The user is asking about a network connection device that can build tables to identify addresses on a network. ... \textcolor{green!60!black}{I will systematically explore possible answers and critically assess the evidence for each.} \textcolor{orange!75!black}{First, could it be a "router"?} ... Does a router specifically "build tables" to identify addresses? ...  \textcolor{red!80!black}{Therefore, the primary function of a router does not seem to align perfectly with the description provided.} \textcolor{orange!75!black}{Next, could it be a "switch"?} ... They use a dynamic or static mapping of MAC addresses to IP addresses. \textcolor{green!60!black}{This does not seem to match the description either.} \textcolor{orange!75!black}{Could it be a "firewall"?} ... After considering these possibilities, ... I have critically examined each candidate and found that \textcolor{orange!75!black}{I lack sufficient evidence to confidently identify the correct answer.} & 
    
    \textbf{Answer: \textcolor{green!60!black}{router}}
    
    \textbf{Reasoning: }To answer the question, I need to ... \textcolor{green!60!black}{The first possibility that comes to mind is a "router".} ... They use routing tables ... \textcolor{green!60!black}{which aligns perfectly} with the description in the question. \textcolor{green!60!black}{However, I should consider other possible devices ...} \textcolor{orange!75!black}{Another possibility could be a "switch".} ... However, switches do not maintain ...  \textcolor{green!60!black}{This makes them less likely to fit the description provided in the question.} \textcolor{orange!75!black}{Another candidate might be a "firewall".} ... This makes them a less likely match. ... \textcolor{green!60!black}{To further evaluate the evidence for each:} - Routers are widely used in ... - Switches are ..., making them less relevant. - Firewalls are ..., making them a weaker fit. \textcolor{green!60!black}{After considering the evidence,} the router stands out as the most accurate answer. \\
    \midrule
    
    \bottomrule
    \end{tabular}
    }
    \caption{Additional Reasoning path examples of different methods on DeepSeek-R1-Distill-Llama-8B. We colored with \textcolor{red!80!black}{wrong factual thinking leading to \textbf{wrong answers}}, \textcolor{orange!75!black}{factual thinking alerting \textbf{not sure}} and the \textcolor{green!60!black}{good factual thinking leading to \textbf{correct answers}}.}
    \label{tab:main_example_appendix}
    \vspace{-3ex}
\end{table}
\begin{table}
    % \small
    \centering
    \resizebox{\linewidth}{!}{
    \begin{tabular}{|p{4.8cm}|p{5cm}|p{5.2cm}|p{5cm}|}
    \toprule
    \multicolumn{4}{c}{\textbf{Reasoning Examples for different models}} \\
    
    \midrule
    \textbf{ICL-IDK} & \textbf{Distill} & \textbf{Ours (SFT)} & \textbf{Ours (GRPO)}\\ 
    \midrule
    
    \multicolumn{4}{|c|}{Question: What do methanogens act as in sewage treatment plants?} \\
    \midrule
    \textbf{Answer: \textcolor{red!80!black}{Biogas producers}} \par \textbf{Reasoning: }Okay, so I need to figure out ... I remember that methanogens are types of bacteria, \textcolor{orange!75!black}{but I'm not exactly sure about their role in sewage} ... I recall something about \textcolor{red!80!black}{biogas being produced} ... So \textcolor{orange!75!black}{perhaps} methanogens convert the organic waste into biogas ... Alternatively, \textcolor{orange!75!black}{maybe} they're acting as \textcolor{green!60!black}{decomposers} ... \textcolor{orange!75!black}{I'm a bit confused} because I also heard that in some cases, methane is released as a byproduct, which can be harmful ... I'm pretty sure that's the case, \textcolor{orange!75!black}{but I'm not 100\% certain.} \textcolor{red!80!black}{I think I'll go with that.} Methanogens in sewage treatment plants act as biogas producers ... & 

    \textbf{Answer: \textcolor{red!80!black}{methane}} \par \textbf{Reasoning: }Okay, so the question is asking ... I remember that \textcolor{green!60!black}{methanogens are involved in breaking down organic matter.} ... \textcolor{orange!75!black}{Maybe methanogens are responsible for converting certain compounds.} ... So the answer could be that they \textcolor{red!80!black}{produce methane gas as part of the anaerobic digestion.} \textcolor{green!60!black}{But the question is phrased as "act as", which might mean their role or function.} ... So \textcolor{orange!75!black}{maybe the answer is that they produce methane gas} ... \textcolor{green!60!black}{Let me check my notes.} ... \textcolor{green!60!black}{Let me confirm.} ... \textcolor{red!80!black}{So the answer is methane.} ...  But the question says "act as", \textcolor{orange!75!black}{so maybe the answer is methane.} ... Yeah, I think that's right. & 
    
    \textbf{Answer: \textcolor{orange!75!black}{I don't know}} \par \textbf{Reasoning: }The user is asking ... First, I consider the possibility ... act as \textcolor{green!60!black}{decomposers} ... \textcolor{green!60!black}{However, what specific evidence supports this?} ... \textcolor{orange!75!black}{But are they exclusively decomposers?} ... I need to investigate further. Next, I consider ... act as aerobes. ... \textcolor{green!60!black}{Do I have any evidence that} ...  \textcolor{orange!75!black}{However, I cannot entirely rule out the possibility that} ... I need to challenge this assumption. Another possibility is ... act as catalysts ... \textcolor{green!60!black}{Do I have any evidence} ... \textcolor{orange!75!black}{I cannot definitively say they act as catalysts} ... Finally, I consider ... act as nitrogen-fixers ... \textcolor{green!60!black}{Is there evidence that} ... \textcolor{orange!75!black}{so this seems less likely.} \textcolor{green!60!black}{After evaluating these possibilities,} I find that ... However, \textcolor{orange!75!black}{I lack definitive evidence to} ... \textcolor{orange!75!black}{I cannot confidently} ... & 
    
    \textbf{Answer: \textcolor{green!60!black}{decomposers}} \par \textbf{Reasoning: }To answer the question ... The first \textcolor{orange!75!black}{possibility} that comes to mind is \textcolor{green!60!black}{decomposers.} ... \textcolor{green!60!black}{I need to verify this} ... Another \textcolor{orange!75!black}{possibility} is reducers. However, ... \textcolor{green!60!black}{This suggests that "reducers" is not the correct answer.} A third \textcolor{orange!75!black}{possibility} is filters. However, ... \textcolor{green!60!black}{This rules out "filters" as a plausible answer.} Next, I need to \textcolor{green!60!black}{evaluate the evidence for "decomposers."} ... This strongly supports the idea ... \textcolor{green!60!black}{To challenge this assumption, I must ask: } ... While ... This suggests that "decomposers" is the most accurate answer. Finally, \textcolor{green!60!black}{I verify this conclusion by considering} ... This confirms that their role as decomposers. \\
    
    \midrule

    \multicolumn{4}{|c|}{Question: What team does controversial quarterback Michael Vick play for in the National Football League in the USA?} \\
    \midrule
    \textbf{Answer: \textcolor{red!80!black}{Cleveland Browns}} \par \textbf{Reasoning: }Okay, so I need to figure out ... \textcolor{orange!75!black}{But I'm not entirely sure about} ... \textcolor{red!80!black}{I think he was with the Cleveland Browns} ... \textcolor{orange!75!black}{But I'm not 100\% sure.} \textcolor{green!60!black}{I should double-check that.} ... I think he might have played for the Cleveland Browns during the early 2000s. Maybe around 2003 or so. ... \textcolor{orange!75!black}{Wait, was it the Cleveland Browns or the Philadelphia Eagles?} \textcolor{red!80!black}{I think he was with the Browns when he got suspended by the Eagles.} So putting it all together, \textcolor{red!80!black}{Michael Vick played for the Cleveland Browns in the NFL.} I think that's the correct answer. & 

    \textbf{Answer: \textcolor{red!80!black}{Miami Dolphins}} \par \textbf{Reasoning: }Okay, so I need to figure out ... I think he started with the Philadelphia Eagles. \textcolor{orange!75!black}{Wait, but then he was traded to the Cleveland Browns, right?} ... After that, he went to the Denver Broncos, and then \textcolor{orange!75!black}{perhaps the Cleveland Browns again? Or maybe ... ?} ... I think the last team he played for was the Miami Dolphins. \textcolor{green!60!black}{But I should confirm.} ... \textcolor{green!60!black}{Wait, another angle: his draft year.} ... \textcolor{red!80!black}{So the answer should be Miami Dolphins.} ... But wait, \textcolor{orange!75!black}{maybe he was traded again after Miami?} I don't recall him ... \textcolor{red!80!black}{So the answer is Miami Dolphins.} & 
    
    \textbf{Answer: \textcolor{green!60!black}{Philadelphia Eagles}} \par \textbf{Reasoning: }To determine ...  The most ... that comes to mind is the \textcolor{green!60!black}{Philadelphia Eagles.} ... However, \textcolor{green!60!black}{I need to verify this and consider other possibilities.} \textcolor{orange!75!black}{Other teams that might be relevant include:} ... \textcolor{green!60!black}{Evaluate evidence for each possibility} 1. \textcolor{orange!75!black}{Miami Dolphins}: ... but there is no evidence to suggest ... 2. \textcolor{orange!75!black}{Denver Broncos}: ... but the Broncos are not the final team he ended up with. 3. \textcolor{orange!75!black}{Philadelphia Eagles}: ... \textcolor{green!60!black}{This is the team he ended up playing for after the trade.} ... \textcolor{green!60!black}{Challenge assumptions.} Could I be misinterpreting the evidence? ... No, the evidence overwhelmingly points to the Eagles ... \textcolor{green!60!black}{Verify the evidence} ... This is consistent with the timeline of his NFL career. & 
    
    \textbf{Answer: \textcolor{green!60!black}{Eagles}} \par \textbf{Reasoning: }To determine ... The \textcolor{orange!75!black}{first possibility} that comes to mind is the \textcolor{green!60!black}{Philadelphia Eagles.} ... but \textcolor{green!60!black}{I need to verify this.} ... \textcolor{orange!75!black}{Another possibility} is the Miami Dolphins ... However, \textcolor{green!60!black}{I don't recall any specific evidence} ... but I need to keep it in mind for further verification. \textcolor{orange!75!black}{A third possibility} is the Dallas Cowboys ... \textcolor{green!60!black}{However, I don't have any specific evidence} ... To resolve this, \textcolor{green!60!black}{I need to focus on concrete evidence.} ... This aligns with my initial thought. Additionally, the Eagles ... Next, \textcolor{green!60!black}{I should challenge this assumption.} ... To verify, I check reliable sources like ...  Finally, \textcolor{green!60!black}{I apply adversarial self-critique.} ... The evidence is sufficient ... \\
    
    \midrule

    \multicolumn{4}{|c|}{Question: Who created Rumpole of the Bailey?} \\
    \midrule
    \textbf{Answer: \textcolor{red!80!black}{Andrew Marquardt}} \par \textbf{Reasoning: }Okay, so I need to figure out ... \textcolor{orange!75!black}{Wait, isn't that by Andrew Marquardt? Or was it someone else?} I'm a bit confused ... \textcolor{orange!75!black}{Maybe it's by David Suchet?} Wait, no, I think he was the actor who played Rumpole. ... Let me think about the name again. The play is by Andrew Marquardt, right? \textcolor{orange!75!black}{But wait, I'm not entirely sure.} \textcolor{green!60!black}{I think I should check} ... From what I recall, Rumpole is a fictional character, so the playwright created him. Therefore, the creator would be Andrew Marquardt. \textcolor{orange!75!black}{I'm a bit unsure} ... but I think in this case, it's the playwright who created the character. \textcolor{red!80!black}{So I'll go with Andrew Marquardt as the creator.} & 

    \textbf{Answer: \textcolor{red!80!black}{Charles Dickens}} \par \textbf{Reasoning: }Okay, so I need to figure out ... I know that Charles Dickens wrote a lot of classic stories. Rumpole of the Bailey... I think that's from "Oliver Twist." \textcolor{orange!75!black}{Wait, no, Rumpole is the character in "Oliver Twist," right?} ... \textcolor{orange!75!black}{But was Rumpole a fictional character created by Dickens?} ... Wait, \textcolor{orange!75!black}{maybe} the answer is Charles Dickens. ... \textcolor{green!60!black}{But I should make sure.} Let me recall. ... \textcolor{red!80!black}{So Dickens created Rumpole of the Bailey.} ... Dickens often created his own characters, so yes, he created Rumpole. \textcolor{green!60!black}{But maybe there's a historical figure with that name? Let me check my memory.} \textcolor{red!80!black}{Charles Dickens wrote "Oliver Twist," and Rumpole is the friend of Oliver.} So the answer should be Charles Dickens. & 
    
    \textbf{Answer: \textcolor{orange!75!black}{I don't know}} \par \textbf{Reasoning: }The user is asking about ... First, I consider whether the creator could be a famous British comedian ... However, I must ask myself: \textcolor{green!60!black}{Do I have any specific recollection or evidence ...?} \textcolor{orange!75!black}{No, I don't have any concrete evidence or memory of} ... Next, I consider whether the creator might be a lesser-known figure ... Again, \textcolor{green!60!black}{I must challenge this assumption:} ... \textcolor{orange!75!black}{No, I don't have any such evidence.} ... Finally, I consider whether the creator might be ... However, \textcolor{green!60!black}{I must ask myself: Do I have any evidence} ... \textcolor{orange!75!black}{No, I don't have any such evidence.} \textcolor{green!60!black}{I have systematically examined multiple possibilities.} ... For each possibility, \textcolor{green!60!black}{I've asked myself what specific evidence I would need to make this claim confidently}, and \textcolor{orange!75!black}{I find that I don't possess such evidence.} & 
    
    \textbf{Answer: \textcolor{orange!75!black}{I don't know}} \par \textbf{Reasoning: }The user is asking about ... First, I need to consider whether the show was created by ... \textcolor{green!60!black}{What evidence supports this?} \textcolor{orange!75!black}{I don't have any direct evidence} ... Next, could it have been created by ... ? \textcolor{orange!75!black}{However, I don't have any specific information} ... \textcolor{green!60!black}{This possibility also seems unsupported.} Could it have been created by a more recent ... ? However, \textcolor{orange!75!black}{I still lack specific evidence} ... \textcolor{green!60!black}{Without concrete evidence, I cannot confidently assert this as the answer.} ... \textcolor{green!60!black}{I have systematically examined multiple possibilities.} ... For each possibility, \textcolor{green!60!black}{I have asked what specific evidence I would need to make this claim confidently}, and \textcolor{orange!75!black}{I find that I don't possess such evidence.} \\
    
    \midrule
    
    \bottomrule
    \end{tabular}
    }
    \caption{Reasoning path examples of different methods on DeepSeek-R1-Distill-Qwen-7B. We colored with \textcolor{red!80!black}{wrong factual thinking leading to \textbf{wrong answers}}, \textcolor{orange!75!black}{factual thinking alerting \textbf{not sure}} and the \textcolor{green!60!black}{good factual thinking leading to \textbf{correct answers}}.}
    \label{tab:main_example_appendix_qwen}
    \vspace{-3ex}
\end{table}

\section{Randomness in GRPO training}
\label{sec:randomness}

\begin{table}[H]
\centering
\label{tab:seed_variability}
\begin{tabular}{lccc}
\toprule
\textbf{Method} & \textbf{Accuracy (Avg.)} & \textbf{Truthfulness (Avg.)} & \textbf{Reliability (Avg.)} \\
\midrule
Distill & 38.43 & 39.33 & 39.33 \\
BARREL (reported in our paper) & 40.70 & 70.40 & 61.58 \\
BARREL (with 5 random seeds) & $41.11 \pm 1.79$ & $69.31 \pm 4.14$ & $61.44 \pm 0.97$ \\
\bottomrule
\end{tabular}
\caption{Seed variability experiments for DeepSeek-R1-Distill-Llama-8B.}
\end{table}

Compared to the Distill baseline, our method achieves notable gains in reliability: 1) DeepSeek-R1-Distill-Llama-8B: from 39.33\% to 61.58\%. 2) DeepSeek-R1-Distill-Qwen-7B: from 26.21\% to 53.12\%. These improvements exceed 20\%, which we believe is substantial and not marginal. But it could be necessary to account for randomness:

\paragraph{Fair Comparison}
We used default random seed settings and made no hard-coded decisions beyond the reward design, ensuring fair conditions across methods.

\paragraph{On Random Variability}
It is widely acknowledged that randomness is inherent in LRM RL training and we could also prove our improvement is statistically significant, as shown by the seed variability experiments in Table~\ref{tab:seed_variability}.

As shown, the standard deviation for reliability is 0.97, which is significantly smaller than the 20\% improvement we report. Thus, the observed performance gains cannot be attributed to randomness.

We also analyzed the randomness of sampling 1k subsets from the test set and found its impact on reliability to be within 1\%. 

\adr{\section{Analysis of Sampling Variance and Stability}
\label{sec:appendix_variance}

To ensure that the performance gains of BARREL are robust and not artifacts of sampling noise, we conducted a detailed analysis of output variance. A key concern in abstention-based systems is whether the decision to abstain is stable across different sampling runs. We investigated the stability of the model's decision-making process—specifically, how often the model fluctuates between answering and abstaining on the same input.

\subsection{Measurement of Inconsistency}
We quantified the variance by measuring the \textit{inconsistency rate} across $k=4$ independent sampling runs for each question in the test set. Since the vanilla model always attempts an answer while BARREL can choose to abstain, we define inconsistency for each setting as follows:

\begin{itemize}
    \item \textbf{Vanilla Model:} Defined as the proportion of cases where the model fluctuates between generating a \textit{correct} and an \textit{incorrect} response. This measures the inherent instability of the base model's knowledge retrieval.
    \item \textbf{BARREL:} Defined as the proportion of cases where the model fluctuates between \textit{answering} and \textit{abstaining}. This measures the stability of our proposed tuning method's decision boundary.
\end{itemize}

\subsection{Quantitative Results}
Table~\ref{tab:variance_comparison} presents the comparison of inconsistency rates across different backbone models. Contrary to the concern that an additional decision module might introduce instability, BARREL consistently exhibits lower variance compared to the vanilla baselines.

\begin{table}[h]
    \centering
    \resizebox{0.8\linewidth}{!}{
    \begin{tabular}{lcc}
        \toprule
        \textbf{Backbone Model} & \textbf{Vanilla Inconsistency} & \textbf{BARREL Inconsistency} \\
        \midrule
        DeepSeek-Distill-Llama-8B & 31.93 & \textbf{19.43} \\
        DeepSeek-Distill-Qwen-7B  & 23.13 & \textbf{17.13} \\
        Qwen3-8B                   & 18.97 & \textbf{14.87} \\
        \bottomrule
    \end{tabular}
    }
    \caption{\adr{Comparison of output inconsistency rates (\%) across 4 independent samples. Lower values indicate higher stability. BARREL demonstrates greater consistency in its decision to abstain than the vanilla model shows in its correctness.}}
    \label{tab:variance_comparison}
\end{table}

For example, on the DeepSeek-Distill-Llama backbone, the inconsistency rate drops significantly from 31.93\% (Vanilla) to 19.43\% (BARREL). This indicates that BARREL does not amplify instability; rather, it stabilizes the output by effectively masking uncertain predictions that are prone to fluctuation in the vanilla model.}

\section{Detailed Examples}

We provide more detailed examples on DeepSeek-R1-Distill-Llama-8B and DeepSeek-R1-Distill-Qwen-7B in Table \ref{tab:main_example_appendix} and Table \ref{tab:main_example_appendix_qwen}. In addition to the two main thinking patterns in current LRMs—second-thought spiraling and last-minute guessing—introduced earlier in this paper, we also observed an additional pattern on DeepSeek-R1-Distill-Qwen-7B that sometimes results in incorrect answers where the model initially proposes an incorrect answer and subsequently engages in multiple rounds of self-checking. However, these self-checks are characterized by expressions of uncertainty and lack of firm commitment. Ultimately, the model still adheres to the original incorrect answer. However, this flawed reasoning pattern can also be effectively transformed into a more reliable and coherent one through our BARREL training framework.

\section{Additional Details}
\label{sec:appendix_addition}

We provide the links and licenses of the datasets and code used in our paper as follows:

\paragraph{Code} We conduct SFT using our own codebase, which is built on top of the Transformers library\footnote{\url{https://github.com/huggingface/transformers}} and DeepSpeed\footnote{\url{https://github.com/deepspeedai/DeepSpeed}}. And we conduct GRPO training using the verl framework \footnote{https://github.com/volcengine/verl}.

\paragraph{Data} We make use of the following publicly available datasets. 
(1) \emph{TriviaQA:} Open-domain question–answering corpus drawn from Wikipedia and the web (Apache 2.0 License)\footnote{\url{https://github.com/mandarjoshi90/triviaqa}};

(2) \emph{SciQ:} 13 679 multiple-choice science questions spanning physics, chemistry, biology, and more (CC BY-NC 3.0 License)\footnote{\url{https://huggingface.co/datasets/allenai/sciq}};

(3) \emph{NQ-Open:} Open-domain variant of Natural Questions covering real Google queries (CC BY-SA 3.0 License)\footnote{\url{https://github.com/efficientqa/nq-open}};

(4) \emph{SimpleQA:} Complex factuality benchmark (MIT License)\footnote{\url{https://github.com/openai/simple-evals}};

(5) \emph{MATH-500:} 500-problem subset of the MATH benchmark for compact maths evaluation (MIT License)\footnote{\url{https://huggingface.co/datasets/HuggingFaceH4/MATH-500}};

(6) \emph{MATH:} full-scale mathematics problem benchmark (MIT License)\footnote{\url{https://github.com/hendrycks/math}}.

(7) \emph{SelfAware:} unanswerable questions (Apache 2.0 License) \footnote{https://github.com/yinzhangyue/SelfAware}.

\section{Models Used in Our Experiments}
We provide the download links to the models used in our experiments as follows:
\begin{itemize}
    \item DeepSeek-R1-Distill-Llama-8B (\url{https://huggingface.co/deepseek-ai/DeepSeek-R1-Distill-Llama-8B})
    \item DeepSeek-R1-Distill-Qwen-7B (\url{https://huggingface.co/deepseek-ai/DeepSeek-R1-Distill-Qwen-7B})
    \item DeepSeek-R1-Distill-Qwen-32B (\url{https://huggingface.co/deepseek-ai/DeepSeek-R1-Distill-Qwen-32B})
    \item DeepSeek-R1-Distill-Llama-70B (\url{https://huggingface.co/deepseek-ai/DeepSeek-R1-Distill-Llama-70B})
    \item DeepSeek-R1 (\url{https://huggingface.co/deepseek-ai/DeepSeek-R1})
    \item QwQ-32B (\url{https://huggingface.co/Qwen/QwQ-32B})
    \item Qwen3-8B (\url{https://huggingface.co/Qwen/Qwen3-8B})
\end{itemize}

\section{Quick Analysis of the Underlying Mechanism} Previous work \citep{liang2024learningtrustfeelingsleveraging} has shown that using the hidden states of LLMs to build a classifier can achieve high consistency with the sampling strategy mentioned in Section \ref{sec:label}. This suggests that the model has the potential to recognize knowledge boundaries and to say "sorry" when faced with unknown questions. In our approach, however, we boost the ability to identify these boundaries through an explicit reasoning process.

\adr{\section{Discussion on Domain Generalization}
\label{sec:discussion_generalization}

While our primary experiments focus on Factual QA, the proposed \textsc{Barrel} framework establishes a principled training paradigm designed for broader applicability. The core methodology—utilizing SFT to seed refusal behaviors and GRPO to instill calibrated abstention incentives—is agnostic to the specific domain. However, adapting this framework requires different patterns of ``reliable reasoning'' tailored to the specific failure modes of different tasks.

\paragraph{Reliability and the Nature of ``Overthinking''.}
In the factual domain, we identify ``overthinking'' (generating extensive rationales for unknown facts) as a primary symptom of unreliability. Consequently, our current implementation specifically targets this behavior. However, this symptom does not apply for all domains. For instance, our preliminary analysis on instruction-following tasks (e.g., IFEval) reveals minimal difference in thought chain length between correct and incorrect responses (e.g., 496 vs. 501 tokens for DeepSeek-R1-Distill-Llama-8B), suggesting that lengthy reasoning is not pathological for reliability issue.

\paragraph{Adapting \textsc{Barrel} to Reasoning Domains.}
Adapting BARREL framework to other domains is still an exciting open question. To generalize \textsc{Barrel} to mathematics or complex instruction following, the construction of the SFT refusal data and the calibration of GRPO rewards should shift to modeling domain-specific boundary patterns:

\begin{itemize}
    \item \textbf{Instruction Following:} Reliability in this domain is tied to constraint satisfaction and execution planning rather than fact retrieval. A reliable reasoning pattern should involve simulating execution steps and proactively detecting potential constraint violations. The SFT phase would thus focus on demonstrating ``simulation-and-check'' behaviors before committing to an answer.
    
    \item \textbf{Mathematics:} Mathematical tasks involve complex logical chains where errors propagate. Here, mitigating unreliability requires sensitivity to intermediate reasoning flaws. A reliable process entails detecting when a logical step becomes uncertain, interrupting the flawed chain, and transitioning to an expression of uncertainty. The GRPO reward modeling would explicitly penalize the completion of hallucinated derivations while rewarding early error detection.
\end{itemize}

In summary, while the specific manifestations of reliable reasoning differ across domains, \textsc{Barrel} provides the foundational infrastructure to verify LRMs' ability to reason reliably. Future work can leverage this framework by formalizing domain-specific reliable reasoning patterns to construct targeted SFT data and calibrated reward signals.}

\section{Limitations}
\label{sec:appendix_limitation}

Although we have evaluated and validated the performance of our method, restricted by our limited computing resource, we mainly utilize DeepSeek-R1-Distill-Llama-8B, DeepSeek-R1-Distill-Qwen-7B and Qwen3-8B to perform our study, which are relatively small LRMs.

In our experiments, though we have covered different datasets, we mainly utilizing verifiable questions, which could be evaluated at test time and reward at training time using string matching method. How to teach LRMs to learn knowledge boundary and behave more deliberatively on open-end questions, like writing articles or providing opinions, remains a valuable topic for future work.

We adopted the general acknowledged sampling strategy to annotate question types, but \adr{we acknowledge that this proxy has inherent limitations. For instance, when the model is genuinely confused between two specific options (e.g., oscillating between distinct answers due to uncertainty), the current consistency-based metric might fail to capture this state accurately,} potentially introducing errors. As a result, there is a certain proportion of mislabeled data in the SFT stage. Our subsequent adaptive GRPO training will help alleviate this problem, but we need to emphasize that \adr{constructing better proxies that can distinguish between ignorance and genuine confusion remains an open challenge, and} it is difficult to obtain a robust model directly through SFT. Detecting issues in the reasoning process remains an open question. If an accurate method for monitoring reasoning could be developed, both false positives and true negatives could be reduced to some extent.

\adr{As shown in Appendix \ref{sec:appendix_variance}, although BARREL improves stability over vanilla baselines, it still exhibits an inconsistency rate of approximately 20\% across sampling runs. This indicates that for a subset of borderline queries, the decision to abstain remains sensitive to sampling noise, suggesting room for further improvement in calibration stability.}

In the paper, we mainly focus on whether the model can reason to claim uncertainty through Reinforcement Learning with Verifiable Rewards (RLVR), without introducing methods such as CoT monitoring and detection. Building detectors for CoT hallucinations and uncertainty is an important future direction, which goes beyond the scope of our current work.

\adr{Another limitation lies in the reward design. Currently, we employ a fixed rejection reward, which implicitly ties the optimal policy to a fixed internal confidence threshold in limited training steps. This could limit the model's flexibility for scenarios requiring different risk tolerances (e.g., higher or lower refusal rates). Future work could explore conditioning the model on a dynamic rejection reward specified in the prompt, enabling the model to generalize to multiple thresholds without retraining.}

\section{Boarder Impact}
\label{sec:appendix_impact}

Although existing LRMs have demonstrated strong capabilities, hallucinations frequently occur in real-world applications, raising doubts about their reliability. Our work enhances the reliability of LLM responses by teaching the models to express uncertainty and refuse to answer when appropriate. \adr{As mentioned in Appendix \ref{sec:discussion_generalization}, future work can leverage BARREL framework by formalizing domain-specific reliable reasoning patterns to construct reliable LRMs for different domains.} We hope our work will inspire more research in this area to further improve model reliability, enabling humans to place greater trust in the positive responses provided by LRMs.

\section{The Use of Large Language Models}
\label{sec:llm_usage}

In this paper, we utilized LLMs to perform grammatical error checking, remove redundancies, and refine the vocabulary in various sections of the paper. It is important to note that the LLM's role was limited to polishing and checking the text, not to writing any paragraphs from scratch.

\end{document}